\documentclass{article}

\usepackage{arxiv}

\usepackage[utf8]{inputenc} 
\usepackage[T1]{fontenc}    
\usepackage{hyperref}       
\usepackage{url}            
\usepackage{booktabs}       
\usepackage{amsfonts}       
\usepackage{nicefrac}       
\usepackage{microtype}      
\usepackage{lipsum}
\usepackage{graphicx}
\graphicspath{ {./images/} }

\usepackage{soul}
\usepackage[super]{natbib}
 \bibpunct[, ]{(}{)}{,}{a}{}{,}%
\usepackage[flushleft]{threeparttable} 
\usepackage{subcaption}
\usepackage{mathtools}
\usepackage{algorithm}
\usepackage{algpseudocode}
\usepackage{enumitem}

\setlist[enumerate]{nosep, leftmargin=1.4cm}
\newlist{steps}{enumerate}{1}
\setlist[steps, 1]{nosep, leftmargin=2cm, label = \textbf{Step\arabic*}}
\usepackage{xcolor}
\usepackage{physics} 
\usepackage{booktabs}
\usepackage{multirow}

\usepackage{hyperref}

\usepackage{setspace}
\makeatletter
\renewcommand\subsubsection{\@startsection{subsubsection}{3}{\z@}%
{-3.25ex\@plus -1ex \@minus -.2ex}%
{1.5ex \@plus .2ex}%
{\normalfont\normalsize\bfseries}}
\makeatother

\usepackage{amsthm,amsfonts,amssymb,amsmath,graphicx}
\usepackage{soul}

\newtheorem{definition}{Definition}

\title{Self-Interest and Systemic Benefits: Emergence of Collective Rationality in Mixed Autonomy Traffic Through Deep Reinforcement Learning}

\author{
Di Chen \\
  Department of Civil and Environmental Engineering\\
  University of California, Davis\\
  Davis, CA 95616 \\
  \texttt{diichen@ucdavis.edu} \\
   \And
 Jia Li \\
  Department of Civil and Environmental Engineering\\
  Washington State University\\
  Pullman, WA 99163 \\
  \texttt{jia.li1@wsu.edu} \\
  \And
  Michael Zhang* \\
  Department of Civil and Environmental Engineering\\
  University of California, Davis\\
  Davis, CA 95616 \\
  \texttt{hmzhang@ucdavis.edu} \\
  * Corresponding author \\
}

\begin{document}
\maketitle
\begin{abstract}
Autonomous vehicles (AVs) are expected to be commercially available in the near future, leading to mixed autonomy traffic consisting of both AVs and human-driven vehicles (HVs). Although numerous studies have shown that AVs can be deployed to benefit the overall traffic system performance by incorporating system-level goals into their decision making, it is not clear whether the benefits still exist when agents act out of self-interest -- a trait common to all driving agents, both human and autonomous. This study aims to understand whether self-interested AVs can bring benefits to all driving agents in mixed autonomy traffic systems. The research is centered on the concept of collective rationality (CR). This concept, originating from game theory and behavioral economics, means that driving agents may cooperate collectively even when pursuing individual interests. Our recent research has proven the existence of CR in an analytical game-theoretical model and empirically in mixed human-driven traffic. In this paper, we demonstrate that CR can be attained among driving agents trained using deep reinforcement learning (DRL) with a simple reward design. We examine the extent to which self-interested traffic agents can achieve CR without directly incorporating system-level objectives. Results show that CR consistently emerges in various scenarios, which indicates the robustness of this property. We also postulate a mechanism to explain the emergence of CR in the microscopic and dynamic environment and verify it based on simulation evidence. This research suggests the possibility of leveraging advanced learning methods (such as federated learning) to achieve collective cooperation among self-interested driving agents in mixed-autonomy systems.

\end{abstract}


\section{Introduction}



Autonomous vehicle (AV) is an emerging technology, and it is anticipated that they will co-exist with human-driven vehicles (HVs) in the foreseeable future. AVs are expected to bring various benefits to mixed autonomy traffic systems, e.g. through coordination and cooperation between vehicles and traffic infrastructures~\citep{yu2019corridor, yang2020eco, typaldos2022optimization}. 


One key assumption often made is that AVs are controlled by one or more shared system-level control objectives, such as system efficiency or energy consumption. However, numerous empirical evidence indicates that individuals tend to be selfish, focusing on maximizing their own utility rather than system-level benefits~\citep{fehr2002altruistic, lange2014social, komorita2019social}. This raises the natural question: when AVs are self-interested and have their own objectives instead of a common one, and there is no enforced external coordination, can cooperation emerge from mixed autonomy traffic intrinsically and can system-level benefits still be achieved? The current literature does not provide a clear answer to this question. The connection between strategic behaviors of AVs and their system level impacts remains elusive~\citep{shang2021impacts, zhong2020influence} and there lacks a good understanding of how cooperation emerges in mixed autonomy traffic, at both individual and collective levels.

To fill this gap, our previous work proposed the concept of collective rationality (CR) for mixed autonomy traffic. Originating from game theory and behavioral economics, CR posits that cooperation can emerge at the collective level even when agents pursue individual interests~\citep{weirich2009collective}. Our earlier works analytically modeled CR as one of the Pareto-efficient equilibria in mixed autonomy traffic~\citep{li2022equilibrium} and empirically verified the existence of CR in real-world mixed traffic~\citep{chen2025evidence}. The primary objective of this paper is to explore the attainability of CR in mixed autonomy traffic through deep reinforcement learning (DRL) with a simple reward design in reproducible microscopic simulation environments. In addition, we aim to postulate and verify a mechanism that explains the emergence of CR.

This research is built on two lines of research in literature. Along the first line, in traffic flow theories, a central problem is understanding how microscopic agent interactions shape the collective behaviors of mixed traffic flow at the macroscopic scale, i.e., the micro-macro connections. Early works focused on the longitudinal behaviors of driving agents, without considering lateral interactions or assuming that lateral interactions are exogenous (i.e., regulated by external lane policies). Examples of longitudinal interactions include the dependency of an AV's car-following mode on the class of the leading vehicles \citep{qin2021lighthill} and the longitudinal control of AV speeds using mean field games \citep{huang2019game}. When considering the lateral dimension, \citet{chen2017towards} and \citet{ghiasi2017mixed} constructed mixed traffic capacity models that accounted for lane policies, such as AV-dedicated lanes. However, the lateral distribution of traffic agents was exogenously specified, unable to capture the agents' endogenous choice.


To capture the endogenous lateral interactions of traffic agents, an early work by \citet{daganzo1997continuum} modeled kinematic waves in a two-lane freeway with a special lane. \citet{daganzo2002behavioral_a, daganzo2002behavioral_b} further explored more complex lateral interactions by modeling macroscopic traffic dynamics in multi-lane freeways, which captures the lane choices between aggressive and timid drivers. The description of discrete lane choices was later generalized to continuous allocation of lateral road spaces. In this context, a division factor, describing the proportion of lateral spaces taken by a class, was introduced to measure lateral interactions between vehicle classes~\citep{logghe2008multi, qian2017modeling}. 
The flow-density relations derived from these models implicitly assume a unique equilibrium. However, what has been overlooked is that the complex interactions of mixed traffic agents can lead to multiple equilibria, some of which arise from the rational decisions of self-interested traffic agents. In light of this, \citet{li2022equilibrium} formally proposed the concept of collective rationality (also known as collective cooperativeness) to model endogenous interactions between mixed traffic agents. The empirical analysis by \citet{chen2025evidence} further verified the existence of CR in real-world mixed traffic of human-driven cars and trucks.




Along another line, research has focused on the pragmatic problem of designing autonomous vehicle behaviors in mixed autonomy environments, with different rationales and purposes. The behavior design problem typically targets either the individual level (micro-scale), aiming to emulate human driving behavior or maximize the vehicle's own utility, or the system level (macro-scale), aiming to improve overall system performance. At individual level, studies typically train AVs to be ``human-like'' or ``utility-maximizing'', employing a wide range of approaches~\citep{driggs2016communicating, driggs2017integrating, yu2018human, sama2020extracting, xu2020learning, hang2020human}. One of the most popular approaches for AV behavior design is DRL, due to its flexibility in incorporating multiple objectives, adaptiveness of controls in dynamic and stochastic environments, and quick online execution. In human-like behavior design, DRL is used to minimize the deviations of AV behaviors from empirical human driving data~\citep{zhu2018human, zhang2018human, wei2018design, shi2021connected, huang2022conditional}. 

On the other hand, extensive evidence from simulations and field experiments indicates that AVs, if their behaviors are not properly designed, may cause negative externalities to the system, including reduced capacity~\citep{calvert2017will, james2019characterizing}, traffic instability~\citep{milanes2014modeling, gunter2019model, gunter2020commercially, shang2021impacts}, and negative impacts on HVs~\citep{zhong2020influence}. Therefore, research explores the idea of using AVs as ``regulators'' in mixed traffic systems, where AV behaviors are designed to optimize system-level metrics such as travel speed, travel time, fuel consumption, safety, and flow stability~\citep{malikopoulos2018optimal, yu2019corridor, huang2019game, goldental2020minority, yang2020eco, jiang2022reinforcement, typaldos2022optimization}. However, these system-level improvements are often achieved at the expense of AVs' own utilities, due to the absence of explicit mechanisms to ensure these individual benefits. Recent literature has attempted to address this issue by adopting locally optimal motion planners and Markov decision processes to minimize fuel and travel time costs for an AV while simultaneously improving the fuel efficiency of surrounding vehicles~\citep{liu2020trajectory, liu2022markov}. However, these methods appear heuristic and lack a theoretical explanation of the underlying mechanisms, as well as guarantees for both individual and system-level performance.

The above discussion suggests the potential of integrating the concept of CR in AV behavior design. Towards this goal, this study examines AV behavior design problem and verifies whether CR can emerge with relaxed assumptions in general dynamic settings. Specifically, we demonstrate that CR can be attained among driving agents trained using deep reinforcement learning with a simple reward design. We examine the extent to which self-interested traffic agents can achieve CR without directly incorporating system-level objectives. A mechanism of CR formation in mixed autonomy traffic is also postulated and verified from simulation evidence.

The major finding and contributions are as follows:
\begin{itemize}
    \item We showed that self-interested traffic agents can attain CR by simple reward design in DRL.
    \item We verified alignment and difference of mixed traffic behaviors with DRL and game-theoretic models.
    \item We proposed and analyzed the underlying mechanism to attain CR in the microscopic and dynamic mixed autonomy environment.
\end{itemize}

The rest of the paper is organized as follows. In Section~\ref{sec:prelim}, we introduce prior definitions of CR and provide a practical example of it in real-world settings. In Section~\ref{sec:DRL_design}, we design the self-interested behaviors of AVs using DRL, and present the experiment setup and model training results. In Section~\ref{sec:attainability}, we examine the attainability of CR in the mixed autonomy traffic with DRL-trained vehicle behaviors. In Section~\ref{sec:mechanism}, we explore the mechanisms that lead to CR in the microscopic and dynamic traffic environment. Finally, we conclude the paper with discussions on major findings and outlooks for future research in Section~\ref{sec:conclusion}.

\section{Prior definition and physical illustration of collective rationality }\label{sec:prelim} 

In this section, we briefly recapitulate the theoretical definition of CR from our earlier game theoretical model~\citep{li2022equilibrium}. Following this, we provide an intuitive example of CR in the physical setting.

\subsection{Collective rationality in equilibrium traffic model}\label{sec:prelim_game} 




In~\citet{li2022equilibrium}, we proposed a two-player bargaining game model to capture the interplay between two classes of driving agents in mixed autonomy traffic, where CR emerges when these interactions converge to certain Pareto-efficient Nash equilibria (NE). In this game, each class of traffic agents is treated as a player. These players are self-interested and perfectly rational, negotiating the road share (i.e., the proportion of lateral spaces) they occupy to maximize driving speeds. Their interactions are modeled as a collective bargaining process. When settling into NE of the bargaining game, the payoff function is written as,
\begin{equation}\label{eq:payoff}
U_i(\rho_1,\rho_2,p_1,p_2) = 
    \begin{cases}
    \begin{aligned}
     &u^*(\rho_1, \rho_2) & \text{if $p_1^*+p_2^* > 1$}, \\
    &u_i(\rho_i/p_i) & \text{if $p_1^*+p_2^* \leq 1$}
    \end{aligned}
    \end{cases}
    i=1,2
\end{equation}
where $\rho_i$ is the traffic density for class $i$, $p_i$ is the bid of road share by class $i$ agent, $u^*(\cdot, \cdot)$ is the one-pipe equilibrium speed, and $u_i(\cdot)$ is the nominal speed function for class $i$. The term $p_i^*$ represents the minimum road share taken by class $i$ at NE, which is computed by $p_i^*=\frac{\rho_i}{u_i^{-1}(u^*)}$. The total minimum road share, $p_1^*+p_2^*$, determines the type of Pareto-efficient equilibrium reached. If $p_1^* + p_2^* > 1$, the 1-pipe equilibrium is Pareto-efficient, where the two classes travel in a fully mixed regime with a synchronized travel speed of $u^*(\rho_1, \rho_2)$. If $p_1^* + p_2^* \leq 1$, the 2-pipe equilibria are Pareto-efficient, with the two classes traveling separately and being better-off than in a fully mixed regime. CR emerges when the mixed traffic settles into 2-pipe NE and is Pareto-efficient. Below, we present the major conclusions regarding the two types of equilibria in the bargaining game model.

\textbf{1-pipe equilibrium}

At one-pipe equilibrium, all traffic agents move at a synchronized speed, denoted as $u^*$. The governing equation for $u^*$ is,
\begin{equation}\label{eq:1pipe_speed}
\frac{1}{\rho_{tot}} 
\left(\frac{\rho_1}{u_1^{-1}(u^*)} \sum_{j=1}^{2} \frac{\rho_j}{a_j} + \frac{\rho_2}{u_2^{-1}(u^*)} \sum_{j=1}^{2} \frac{\rho_j}{b_j}\right) = 1
\end{equation}
where $\rho_{tot}=\rho_1+\rho_2$ is the total system density. The terms $a_j$ and $b_j$ are the scaling parameters for class 1 and class 2, respectively. These scaling parameters capture the ``type-sensitivity'' in the mixed traffic, reflecting how a vehicle class's desired headways or spacings are influenced by the class of the leading vehicle.
The 1-pipe equilibrium speed $u^*$ is computed by solving the implicit function (\ref{eq:1pipe_speed}), where other parameters and functions can be either assumed or calibrated from data~\citep{chen2025evidence}. 
This implicit function follows the refinement by~\citet{chen2025evidence}, where each class is endowed with a distinct nominal speed function $u_i(\cdot)$. The original formulation assumed a single nominal speed function shared by both classes~\citep{li2022equilibrium}.

\textbf{2-pipe equilibria}

When the 2-pipe equilibria are Pareto-efficient, the condition $p_1^* + p_2^* \leq 1$ holds. Given that the total road share is 1, this implies that after each class occupies its minimum road share $p_i^*$, there is a remaining road share. This remaining road share is referred to as the cooperation surplus (also known as the road share surplus). The formal definition is presented in Definition~\ref{def:surplus}. A positive cooperation surplus ($s>0$) is a necessary condition for attaining CR. 
\begin{definition}[Cooperation surplus]\label{def:surplus}
We call the road share left from players' collective bargaining as the cooperation surplus, $s \coloneqq 1 - p_1^* - p_2^*$.
\end{definition}

Furthermore, unlike the 1-pipe Pareto-efficient NE, which is unique, the 2-pipe Pareto-efficient Nash equilibria are non-unique because there are multiple ways to split the cooperation surplus between the two classes. To further characterize CR, \citet{li2022equilibrium} examined how the cooperation surplus is divided between the two classes. This division is quantified by the surplus split factor, formally defined as,
\begin{definition}[Surplus split factor]\label{def:surplus_split}
The effective split of the cooperation surplus by the two players is called the surplus split factor, denoted as $\lambda(\rho_1, \rho_2)$, and $\lambda \in [0, 1]$.
\end{definition}

By this definition, the road share allocated to each class becomes,
\begin{equation}
\begin{cases}
\begin{aligned}
    &p_1 = p_1^* +\lambda(\rho_1, \rho_2) s \\
    &p_2 = p_2^* +(1-\lambda(\rho_1, \rho_2)) s
\end{aligned}
\end{cases}
\end{equation}
where the total road share satisfies $p_1+p_2=1$. The surplus split factor can be empirically estimated when data is available, providing a quantitative measure of the fairness in benefit allocation between two classes of traffic agents~\citep{chen2025evidence}.

\subsection{Example of collective rationality}

\begin{figure}[ht] 
    \centering
    \includegraphics[width=\textwidth]{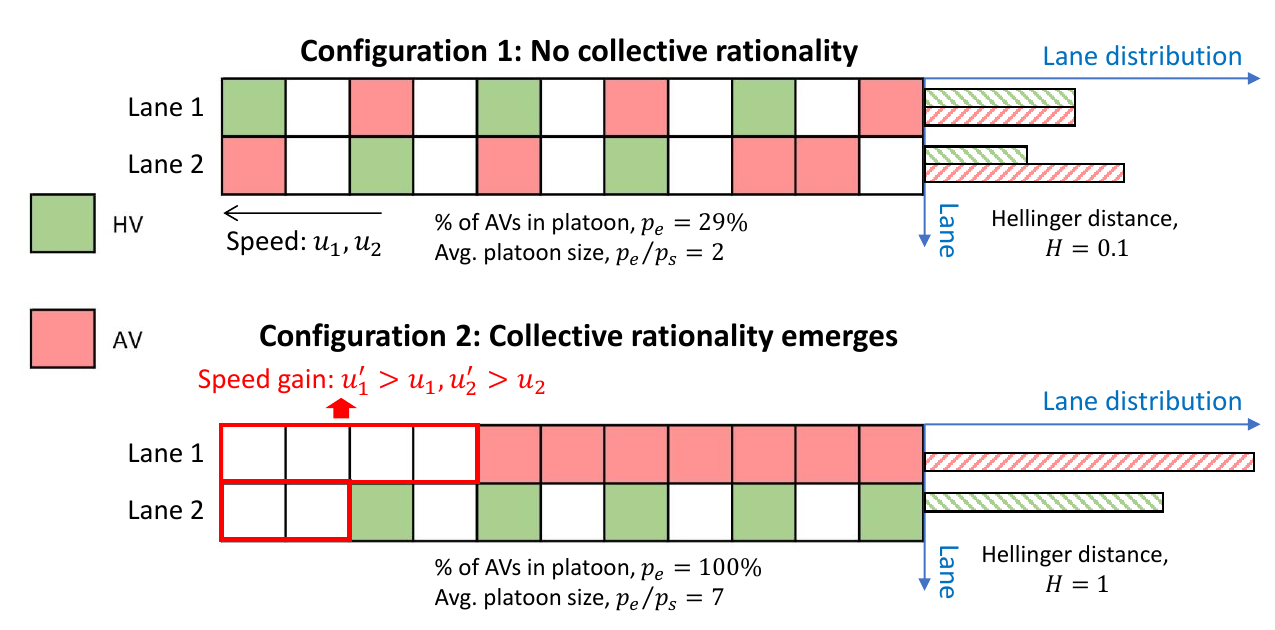}
    \caption{A simple grid illustration of collective rationality in physical world.} 
    \label{fig:conceptual}
\end{figure}

We provide an illustrative example of CR in physical world. A cell representation in Figure~\ref{fig:conceptual} is used to visually compare the scenarios without and with CR. Type sensitivity is assumed, with AVs having a smaller desired time headway than HVs. In a fully mixed configuration (configuration 1), self-interested traffic agents utilize all available road space and travel at a synchronized speed. In contrast, when CR is present (configuration 2), these traffic agents demonstrate a degree of spatial organization that results in increased spacing. This additional spacing enables both classes to achieve higher traffic speeds compared to the fully mixed scenario (i.e., $u_i'>u_i$). In other words, configuration 2 leads to a Pareto improvement over configuration 1, demonstrating the existence of CR.

Furthermore, we offer a preliminary postulation regarding the mechanism behind CR attainment using this example. The grid representations of the two configurations intuitively suggest that improved spatial organization is a key factor in achieving CR. This insight is further supported by the proposed metrics, namely, $p_e$, $p_s$, and $H$, which quantify intra-class platooning and inter-class separation in both the longitudinal and lateral dimensions. Essentially, the emergence of CR in configuration 2 is reflected in higher values of these metrics. Building on this intuition, in later sections, we will propose a mechanism for CR attainment and validate it using simulation evidence. The metrics presented in this figure capture the macroscopic perspective, specifically the spatial organization. To provide a comprehensive understanding of the underlying mechanisms, additional analyses will be conducted at the microscopic and mesoscopic levels.


\section{Decision-making of self-interested autonomous vehicles}\label{sec:DRL_design}

The bargaining game model in \citet{li2022equilibrium} demonstrates the existence of CR in idealized traffic scenarios. This model provides a theoretical benchmark for exploring more realistic settings. However, CR's attainability in more complicated and realistic scenarios cannot be directly concluded from this theoretical model. This is because, as a one-shot, macroscopic analytical model, it does not account for the detailed and evolving interactions of traffic agents in a microscopic environment. The model also overlooks the complexities of real-world driving, such as the presence of physical vehicles and longitudinal car-following dynamics. As a result, it lacks insights into how these interactions evolve over time and whether they can lead to CR in realistic driving scenarios. 

This study aims to explore the emergence of CR in more complex and realistic scenarios. Our goal is not to align these models with the idealized game theoretical model or design traffic agent behaviors that exactly match the theoretical model's players. Instead, we hypothesize, from an analogous perspective, that CR may also emerge in more realistic mixed traffic environments with self-interested traffic agents. To verify this hypothesis, we will design self-interested deep reinforcement learning (SI-DRL) agent behaviors in a mixed autonomy traffic simulation environment, where AVs are treated as learning agents. These SI-DRL agents' decisions will be tailored using a simple reward design. The DRL framework enables SI-DRL agents to adaptively learn and optimize their driving strategies through continuous interaction with the dynamic traffic environment.


\subsection{Preliminaries of deep reinforcement learning}


The foundation for reinforcement learning (RL) is the Markov decision process (MDP), where the two objects, the RL agent (i.e., the decision making entity) and the environment, continuously interact. The MDP involves four major elements: state, action, policy, and reward. The agent at state $s$ in the state space $S$ chooses an action $a$ from the action space $A$ based on the policy $\pi$, which dictates the probability $\pi(s,a)$ of taking the action $a$ at state $s$. The action yields an immediate reward $R(s,a)$, and the agent transitions to a new state, continuing this cycle until a terminal state is reached.
Under a given policy $\pi$, the state value function and action value function are respectively expressed as:
\begin{equation}
    V^{\pi}(s) := E_{\pi} \{ R_t | s_t=s \} = E_{\pi} \biggl\{ \sum_{k=0}^{\infty} \gamma^k r_{t+k+1} | s_t = s \biggr\},
\end{equation}

\begin{equation}\label{eq:mdp_q_value}
    Q^{\pi}(s,a) =  E_{\pi} \{ R_t | s_t=s, a_t=a \} = E_{\pi} \biggl\{ \sum_{k=0}^{\infty} \gamma^k r_{t+k+1} | s_t = s, a_t = a \biggr\}.
\end{equation}

The ultimate objective is to discover an optimal policy $\pi^*$ that maximizes the discounted cumulative reward. To achieve this, we employ the Q-learning algorithm, a model-free method that iteratively updates the action-value function without requiring a model of the environment \citep{watkins1992q}. The Q-learning update rule is given by,
\begin{equation}
    Q(s_t,a_t) \leftarrow Q(s_t,a_t) + \alpha \left( r_{t+1} + \gamma \max \limits_{a} Q(s_{t+1},a) - Q(s_t,a_t) \right)
\end{equation}
where $\alpha$ is the learning rate and $\gamma$ is the discount factor.
In practice, large or continuous state spaces often exceed the capabilities of traditional Q-value tables. To address this, we use a deep Q-network (DQN) that approximates $Q(s,a)$ with an artificial neural network. 

\subsection{Behavior design of self-interested autonomous vehicles}



The SI-DRL agents proposed in this study refers to a class of self-interested AVs that make lane change decisions to maximize their individual driving speed, without incorporating with any system-level objectives. These SI-DRL agents have limited perception, being aware only of their local environment without knowledge of the overall system states. Furthermore, we implement a decentralized training scheme, where there is no external coordination among AVs, and each agent independently optimizes its own reward. This study explores whether these self-interested AVs can achieve CR through their continuous interactions with HVs in the mixed autonomy traffic system.

\begin{figure}[ht]
    \centering
    \includegraphics[width=.6\linewidth]{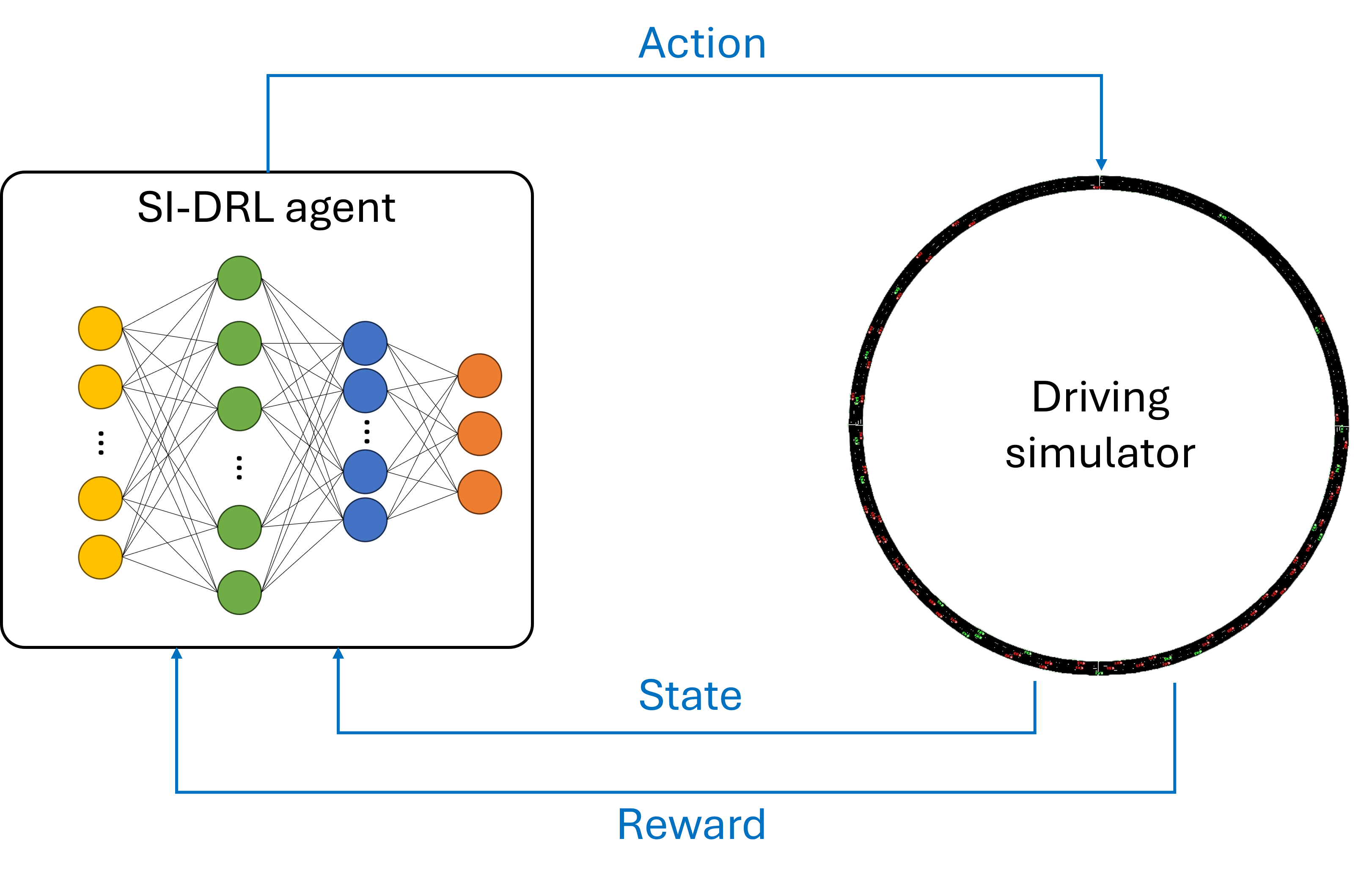} 
    \caption{Deep reinforcement learning in the mixed autonomy simulation environment.} 
    \label{fig:mdp}
\end{figure}

Figure~\ref{fig:mdp} illustrates the decision-making process of SI-DRL agents using DRL in a simulation environment. It incorporates a state representation, a set of actions, and a reward function to guide the learning process.

\subsubsection{State}

\begin{figure}[ht]
    \centering
    \includegraphics[width=.65\linewidth]{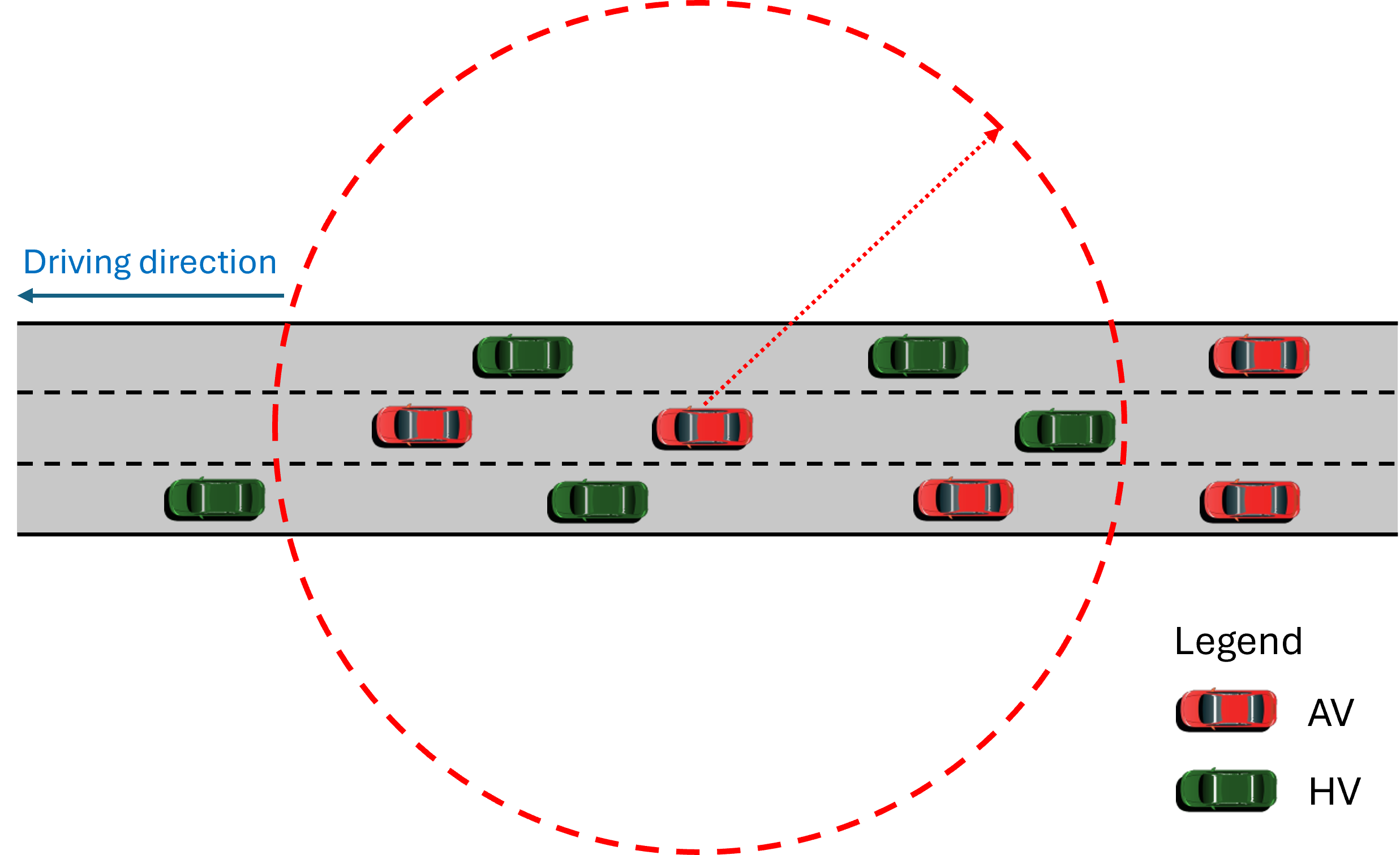} 
    \caption{Sensing range of an autonomous vehicle.} 
    \label{fig:sensing_range}
\end{figure}

An SI-DRL AV's lane change decision is influenced by itself and its interactions with surrounding vehicles within its sensing range. For this study, we assume an AV's sensing range to be a circle with a 100-meter radius, as depicted in Figure~\ref{fig:sensing_range}. 

The state space is composed of two components. The first component, denoted as $X_i^{AV}$, represents the state of the $i$th SI-DRL agent itself. The second component, denoted as $X_i^{surr}$, captures the state of the surrounding vehicles as perceived by the SI-DRL agent. It is represented as,
\begin{equation}
     S_i = \{X_i^{AV}, X_i^{surr} \} 
\end{equation}
where 
\begin{equation}
    X_i^{AV} = \{x_i, y_i, u_i, l_i\}
\end{equation}
\begin{equation}
    X_i^{surr} = \{x_j^{surr}, y_j^{surr}, u_j^{surr}, l_j^{surr}, c_j^{surr} \}, j=1,\dots, n
\end{equation}
Here, $x,y$ represents the coordinates of a vehicle, $u$ represents vehicle speed, and $l$ is an integer representing the lane number a vehicle is traveling at. The variable $c_j^{surr}$ represents the vehicle type (AV or HV) of the $j$th surrounding vehicle, and $n$ quantifies the total number of surrounding vehicles within an AV's sensing range. 

\subsubsection{Action and reward}

The action space of the $i$th SI-DRL agent, denoted by $A_i$, consists of a discrete set of actions representing the agent's lane change decisions, which is written as,
\begin{equation}
    A_i = \{\text{change to left, change to right, keep lane} \}
\end{equation}

The reward function measures the effectiveness of a chosen action in a specific state and is vital in shaping agent behaviors. Since the SI-DRL agent prioritizes increasing individual travel speed, this study formulates a simple reward function to reflect this goal while also considering the external impacts of lane changes on the surrounding traffic. Specifically, the reward function includes two components: the agent's speed, which reflects its self-interest, and a lane change penalty, which discourages frequent lane changes. The reward function is structured as,
\begin{equation}\label{eq:reward_speed}
    R_i = w_{speed} u_i - w_{lane \ change} I_i
\end{equation}
where $I_i$ is a binary variable indicating whether the agent changes lane or not.


\subsection{Experiment setup}

\bigskip
(i) Traffic scenarios and simulation setup
\bigskip
We conduct experiments on a three-lane ring road with a total length of 1000 meters. We use an open-source simulator named Simulation of Urban MObility (SUMO)~\citep{krajzewicz2012recent}, a micro-traffic simulation software capable of modeling multiple agents in an urban-scale environment, to create the ring road configuration and train the DRL models. The SUMO simulator utilizes a traffic control interface (TraCI), which supports retrieving values from simulated objects and manipulating their behaviors. 

To facilitate the study of CR, we examine a wide range of traffic densities and AV penetration rates. We consider four traffic densities: 25 vehicles per mile per lane (vpm/lane), 40 vpm/lane, 55 vpm/lane, and 70 vpm/lane (equivalent to 15 veh/km, 25 veh/km, 34 veh/km and 43 veh/km, respectively). Additionally, we evaluate three AV penetration rates: 25\%, 50\%, and 75\%. This setup results in a total of 12 unique traffic scenario settings. Scenarios corresponding to the fully congested regime are excluded, as their Pareto-efficient NE are more likely to fall within the 1-pipe regime, where CR does not exist~\citep{li2022equilibrium}. 

Each traffic scenario is input to the simulation experiment respectively, with a 100-second warm-up period to ensure that all vehicles are loaded before the DRL training begins.
Departure positions and lane numbers are randomly assigned to each vehicle when loading them into the ring road. 
Additionally, all vehicles are uniformly $5 \ m$ in length and have maximum acceleration and deceleration rates of $2.6 \ m/s^2$ and $4.5\ m/s^2$, respectively, following the default settings by SUMO. Furthermore, to introduce heterogeneity into the simulation environment, HVs are assigned a random maximum speed within the range of $[17.5,\ 25]$ m/s, and AVs within $[21,\ 30]$ m/s.

The lane-changing behaviors of AVs are controlled by the DRL model, while HVs are controlled by SUMO’s built-in dynamics model, LC2013~\citep{erdmann2015sumo}. Besides, the car-following behaviors of AVs and HVs are both governed by the Intelligent Driver Model (IDM)~\citep{treiber2013traffic}.

\bigskip
(ii) Baseline scenario
\bigskip

We define the baseline as a no-control scenario, where there is no DRL control over AVs' lane change decisions. In this case, both lane-changing and car-following maneuvers for AVs and HVs are managed by SUMO's LC2013 and IDM models, respectively. In this context, CR refers to achieving a Pareto-efficient NE under DRL control, relative to the no-control scenario.


\bigskip
(iii) Type-sensitivity parameters
\bigskip

The type-sensitivity between vehicle classes refers to the different desired headways or spacings a class of vehicles maintains based on the class of the leading vehicles. As indicated in Figure~\ref{fig:conceptual}, it is a crucial property for the attainability of CR in mixed traffic, which has been analytically modeled~\citep{li2022equilibrium} and empirically validated in real-world scenarios~\citep{chen2025evidence}. In this study, we assume the presence of type-sensitivity in mixed autonomy traffic, particularly in relation to a vehicle's minimum desired time headway. Due to the lack of empirical data for AVs with current technology, the desired time headways for AVs are primarily based on assumptions~\citep{jiang2023traffic, guo2023freeway}. In this study, by denoting HV as class 1 and AV as class 2, we assume the following minimum headway values: $h_{11}^* = 1.5 \ s$ for a HV following another HV, $h_{12}^*=1.5 \ s$ for a HV following an AV, $h_{22}^*=1 \ s$ for an AV following another AV, and $h_{21}^*=1.25 \ s$ for an AV following a HV. These values suggest that HVs are type-insensitive to the class of their leading vehicles, whereas AVs exhibit type-sensitivity, tending to follow more closely when their leaders are also AVs, due to their communication capabilities.


\bigskip
(iv) Deep-Q network (DQN) and hyperparameters
\bigskip

It is assumed that the SI-DRL AVs belong to a same company, allowing them to utilize a shared DQN for decision-making, thereby reducing computational complexity.
The input layer of the DQN processes the state representation, which aligns with the dimensions of the state space. We set the state space size to $N_s=200$. The DQN architecture includes three fully connected hidden layers, containing 256, 128, and 64 neurons, respectively. Each hidden layer uses a rectified linear unit (ReLU) activation function to introduce non-linearity into the model. The output layer has a number of neurons equal to the number of actions, which is 3 in our case.

To improve the training stability and sample efficiency, we uses experience replay when updating the online network~\citep{lin1992self}. The trajectories $(s_t, a_t, r_{t+1}, s_{t+1})$ generated by $Q_{online}$ are stored in memory buffer $M$ and are randomly sampled to update the online network. This random sampling of trajectories collected across training episodes helps mitigate the temporal correlation of the data. Additionally, to balance the trade-off between action exploration and exploitation, an $\epsilon$-greedy decay function is applied, defined as $p(a) = \epsilon_{end} + (\epsilon_{start}-\epsilon_{end}) e^{-epi/r_{decay}} $. Here, $\epsilon_{start}$ and $\epsilon_{end}$ are the initial and final values of $\epsilon$ respectively, $epi$ is the current episode number, and $r_{decay}$ is the decay rate. We set $\epsilon_{start}=1, \epsilon_{end}=0.01$ and $ r_{decay}=300$. For the optimization, we use the Adam optimizer \citep{kingma2014adam}, with an exponential learning rate decay strategy starting at $10^{-3}$ and a decay factor of 0.99.


\subsection{Model convergence}

\begin{figure}[ht] 
    \centering
    \includegraphics[width=.8\textwidth]{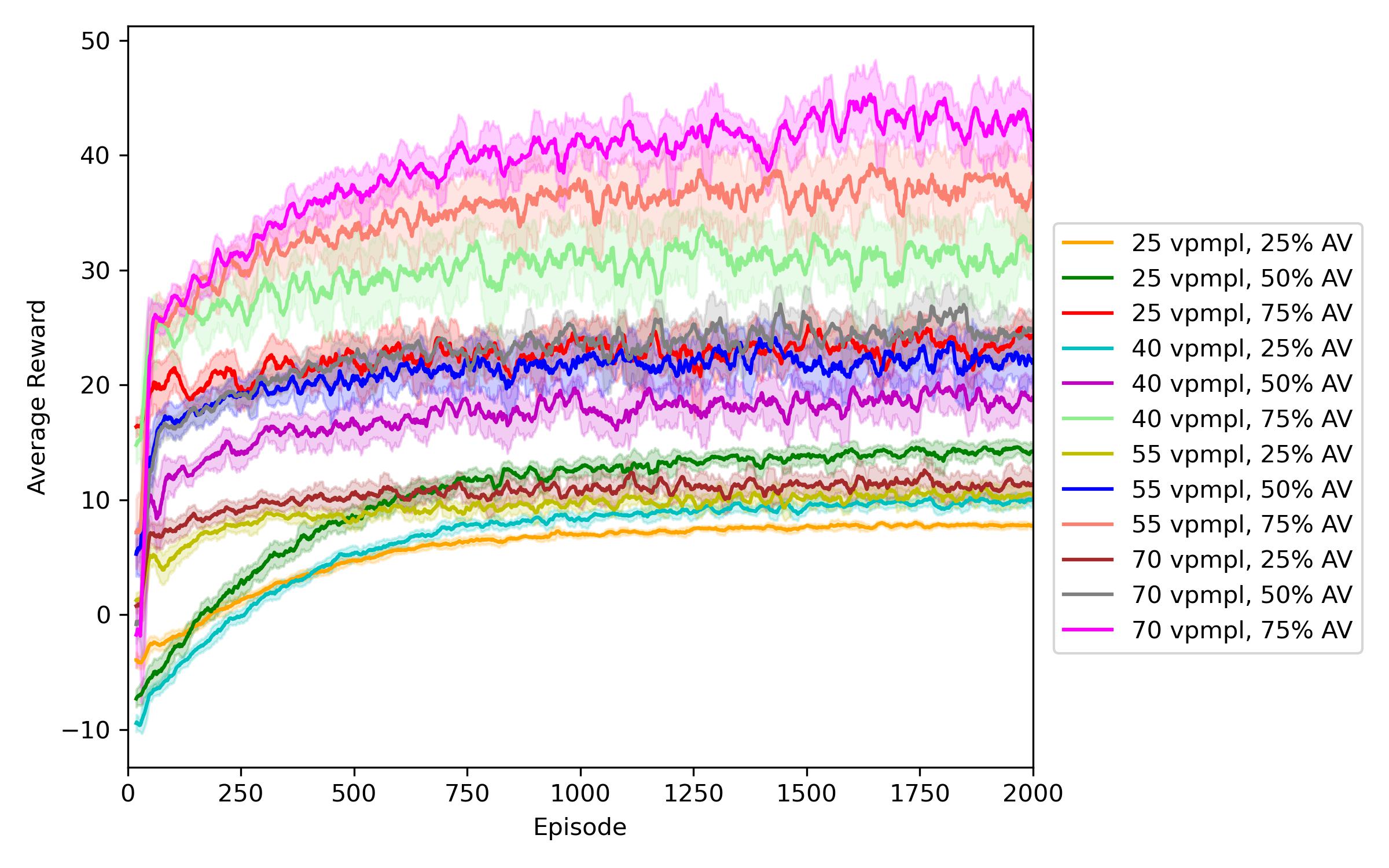}
    \caption{DRL model reward curves for different traffic scenarios.}
    \label{fig:reward}
\end{figure}

We perform 2000 training episodes for each traffic scenario, with the first 30 episodes as a warm-up phase during which SI-DRL agents perform random actions for exploration without any model updates. Figure~\ref{fig:reward} presents the reward curves and their variations, with a smoothing average applied every 20 episodes. The differences in reward scales across traffic scenarios arise from varying traffic conditions and the number of AVs in each scenario. Generally, the DRL model for each scenario converges after approximately 1500 episodes. 

\subsection{Model evaluation and traffic measurements}

We will evaluate the resulting behaviors in mixed autonomy traffic using these converged models. Specifically, 10 evaluation episodes are run for each traffic density and AV penetration rate setting. Each episode lasts 1200 seconds, with the first 100 seconds designated for vehicle loading time. Besides, each episode uses a unique random seed, different from both each other and those used during training. This approach guarantees that the observed behaviors are not artifacts of seed-specific conditions and ensures the reliability of the proposed DRL models in dynamic and unpredictable environments. The data from the DRL evaluation models will be utilized for the subsequent analysis.

Additionally, to facilitate subsequent analysis, we need to obtain macroscopic traffic measurement data from the simulation. For this purpose, we adopt Edie’s generalized definitions of traffic flow characteristics, including traffic flow, density, and speed~\citep{edie1963discussion}. These definitions are based on a space-time region of interest ($\Delta x \times \Delta t$), where vehicle trajectories are present. The traffic flow characteristics are estimated by averaging vehicle trajectories over these space and time domains. To determine the space-time region of this study, we divide the ring road into 4 subsections, each with a length $\Delta x=250 \ m$. And we choose $\Delta t = 8 \ sec$ for the space-time region.

\section{Attainability of collective rationality}\label{sec:attainability}


In this section, we will examine the attainability of CR in mixed autonomy traffic where SI-DRL AV behaviors have been sufficiently trained. We will assess it from two perspectives: first, the attainability of Pareto-efficient NE (Section~\ref{sec:pareto}); second, comparisons of various macroscopic traffic characteristics between the DRL simulation output and the game theoretical model predictions (Section~\ref{sec:compare_game}).


\subsection{Pareto-efficient equilibria}\label{sec:pareto}



We establish criteria to identify equilibrium conditions in the microscopic and dynamic environment, including variations in vehicle speeds and spacings. We first standardize vehicle speeds and spacings to a mean value of 1 to account for varying scales at different density levels, allowing their standard deviations to reflect variability relative to the mean. Based on this, the criteria to determine the time of reaching NE are defined as follows:
\begin{itemize}
    \item Speed variation is less than 0.2 m/s. 
    \item Spacing variation is less than 0.35 m. 
    \item The above criteria maintains for at least 20 seconds. 
\end{itemize}
We note that this definition is a relaxed criterion, where achieving NE does not mean that vehicles will stop perform car-following and lane-changing maneuvers. Instead, vehicles will continue these maneuvers more smoothly to meet their speed gain objectives. In the subsequent analysis, we will sample evaluation data within a 600-second time window after NE is reached to assess the attainability of CR.

According to the theoretical definition, CR is achieved when a Pareto-efficient NE is attained~\citep{li2022equilibrium}. This means that the average speeds for each class of vehicles under DRL control are at least as high as the corresponding vehicle class speeds in the no-control scenario. Unlike the idealized one-shot bargaining game, we acknowledge that perfect Pareto-efficient NE might not be fully attainable in the simulation experiment. This stems from the existence of non-equilibrium behaviors and the stochastic and random nature of traffic dynamics, both of which contribute to the difficulty of accurately measuring traffic regimes. 
Therefore, we introduce a tolerance window $\tau \in [0,1]$, where a traffic scenario is considered Pareto-efficient if the vehicle speed under DRL control is no less than $\tau$ times the speed in the no-control scenario. A smaller $\tau$ value indicates closer proximity to perfect Pareto efficiency, with $\tau=0$ representing the ideal Pareto-efficient NE. We expect the tolerance window to be reasonably small to ensure minimal compromise.

\begin{figure}[ht]
    \centering
    \includegraphics[width=\textwidth]{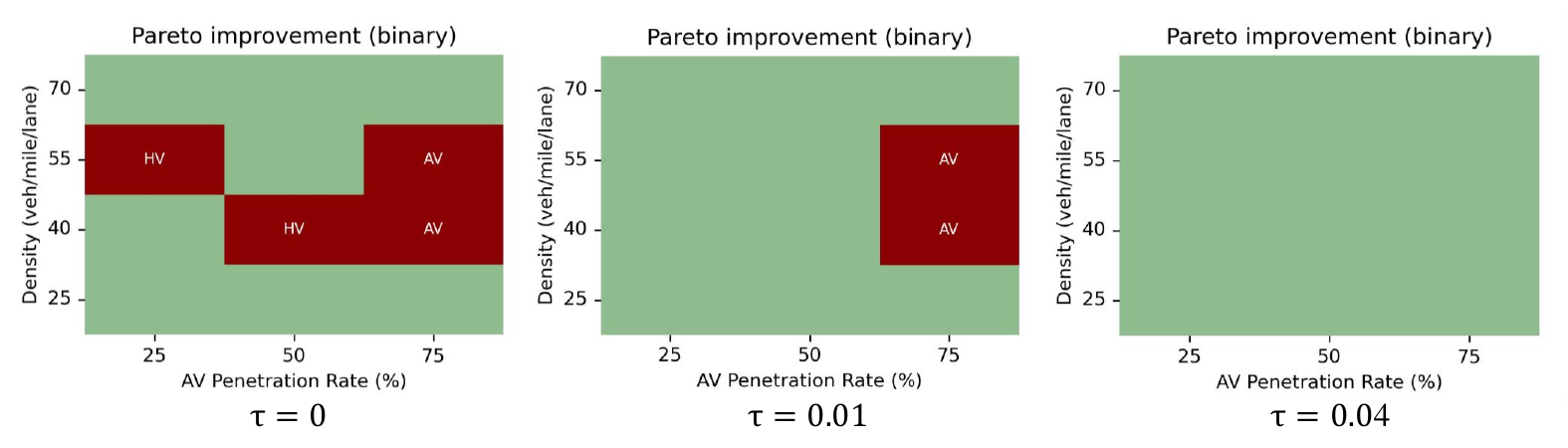}
    \caption{Attainment of Pareto efficiency under DRL control relative to the no-control scenario across different tolerance windows. Green indicates scenarios where Pareto-efficient NE is achieved, while red indicates where it is not. The labels ``HV'' or ``AV'' identify the vehicle class that has not reached Pareto-efficient NE in the corresponding scenario.}
    \label{fig:Pareto}
\end{figure}

Figure~\ref{fig:Pareto} illustrates the attainment of Pareto-efficient NE across traffic scenarios under different tolerance windows. The results indicate that HVs are more likely to achieve Pareto efficiency compared to AVs, requiring only a 1\% tolerance window. This 1\% tolerance window corresponds to a speed reduction of up to 0.15 mph compared to the no-control scenario, which is considered negligible.
Furthermore, the right-most figure demonstrates that a Pareto-efficient NE is achievable when class-level average speeds are allowed to be 4\% lower than those in the no-control scenario. This equates to a maximum speed reduction of 1.5 mph. This minor speed reduction can be attributed to the slight heterogeneity introduced by the external control (i.e., DRL) in the closed-loop ring road system and is thus considered an acceptable trade-off. Given this, Pareto-efficient NE is considered to be achieved across these scenarios.

\subsection{Comparisons with the bargaining game model}\label{sec:compare_game}

In addition to achieving Pareto-efficient NE, observing alignments with the game theoretical model can further support the attainability of CR through DRL. To examine the existence of such alignments, this section compares various macroscopic traffic variables between the DRL evaluation output and the theoretical model's predictions. We note that exact matches are neither expected nor sought due to the inherent randomness in DRL and the simplified assumptions in the theoretical model. Instead, our primary goal is to identify any existing alignments. We will first establish a benchmark game-theoretical model that reaches a Pareto-efficient equilibrium. Then, comparisons will be made from three perspectives: first-order, second-order, and fundamental diagram comparisons.


\subsubsection{Establishment of benchmark game theoretical model}


According to \citet{li2022equilibrium}, there are numerous Pareto-efficient equilibria (i.e., CR) in the 2-pipe regime, determined by various ways of splitting the cooperation surplus. It is essential to properly establish a benchmark bargaining game model that aligns with the Pareto-efficient equilibrium achieved in DRL. This involves determining relevant parameters in the theoretical model based on simulation data, which has been shown to be feasible~\citep{chen2025evidence}. Specifically, (i) the nominal behaviors of players will be defined according to those used in the simulation, and (ii) the optimal surplus split factor will be estimated from the simulation. 

\bigskip
(i) Nominal behaviors of players
\bigskip

\begin{figure}[ht] 
    \centering
    \begin{subfigure}{.49\textwidth}
      \centering
      \includegraphics[width=1\linewidth]{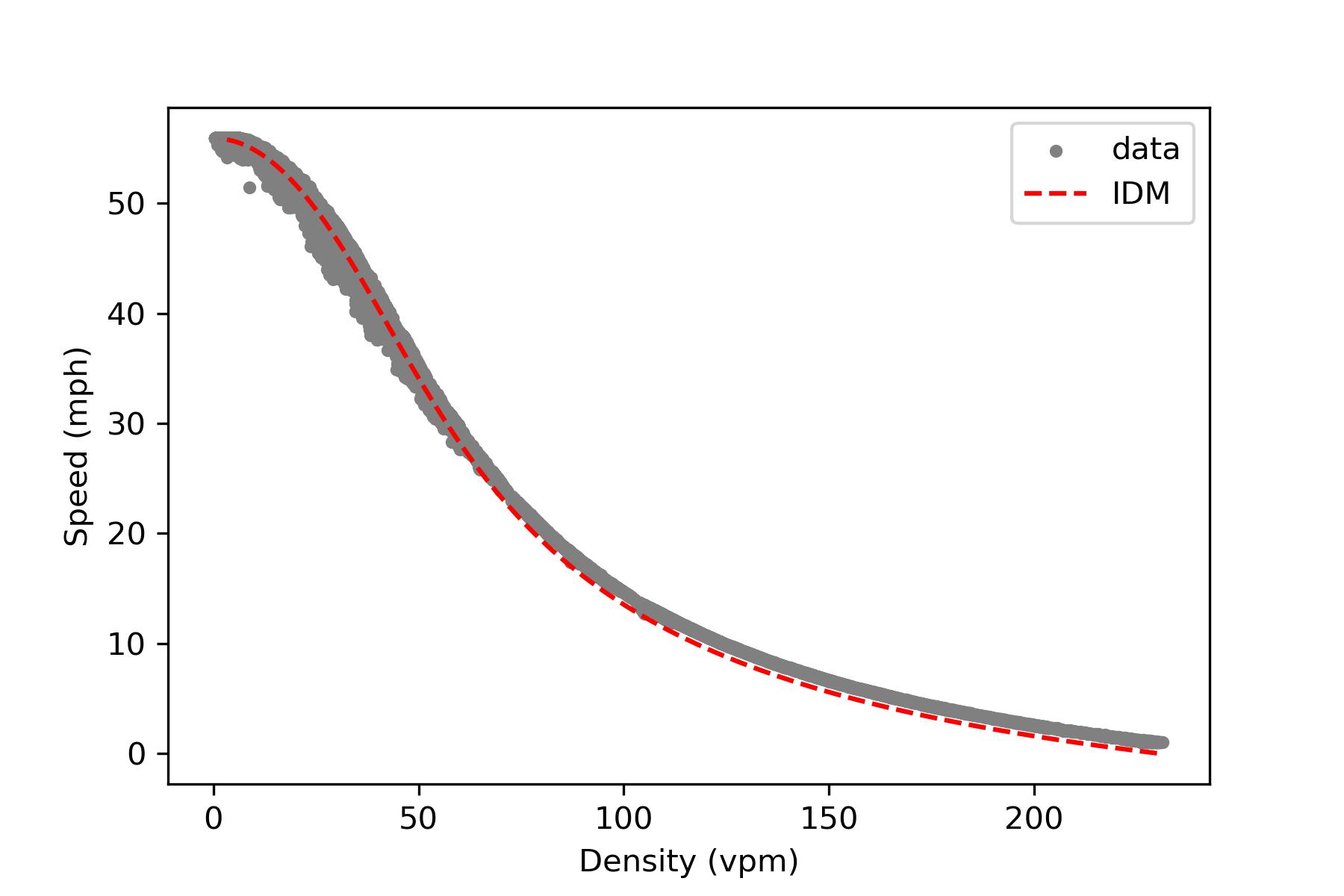}
      \caption{HV (Class 1)}
    \end{subfigure}
    \begin{subfigure}{.49\textwidth}
      \centering
      \includegraphics[width=1\linewidth]{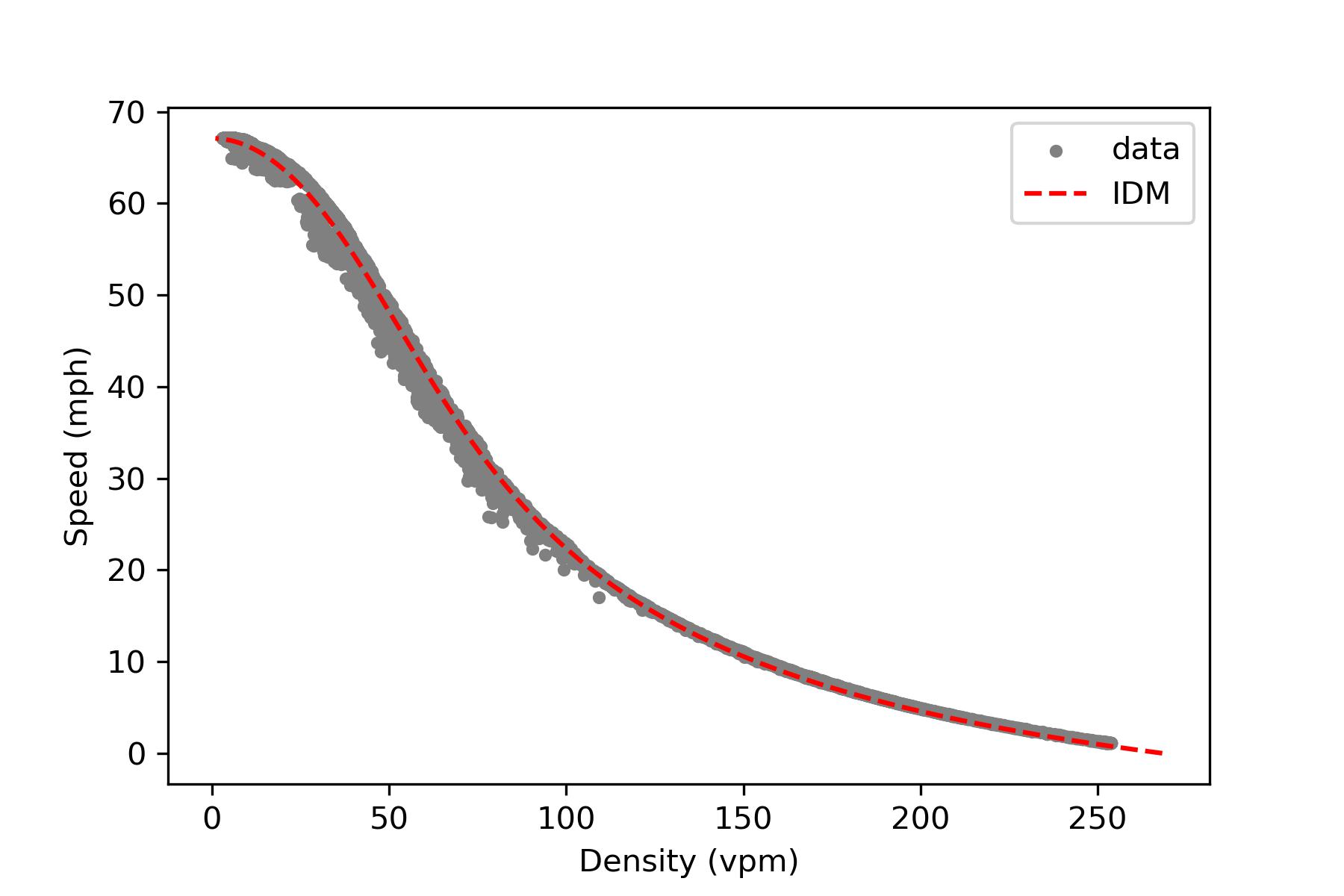}
      \caption{AV (Class 2)}
    \end{subfigure}
    \caption{Comparison of speed-density relations between simulation data and the intelligent driver model (IDM).} 
    \label{fig:u_k_IDM}
\end{figure}

In the pay-off functions (\ref{eq:payoff}), each player is endowed with a nominal speed-density function, $u_i(\cdot)$. In the DRL simulation environment, we use the IDM model for both AVs and HVs, with specific IDM model parameters for each class. To match this setup in the benchmark model, we derive the corresponding speed-density functions of the IDM model under equilibrium conditions and use the same model parameters as in the simulation. The derived functions and the simulation data are plotted in Figure~\ref{fig:u_k_IDM}, showing a close match for both classes. This validates the feasibility of using the IDM model to represent nominal behaviors in the benchmark game theoretical model.

Additionally, we set the scaling parameters introduced in (\ref{eq:1pipe_speed}) to be $a_{1}=1, a_{2}=1, b_{1}=0.8, b_{2}=1$. This setting suggests that class 1 vehicles (HVs) are not sensitive to the class of their leader, whereas class 2 vehicles (AVs) tend to follow AVs more closely than HVs. This setup is consistent with the minimum time headway parameters used in the DRL simulation settings.


\bigskip
(ii) Optimal surplus split factor
\bigskip


According to \citet{li2022equilibrium}, the way to split the cooperation surplus is denoted as $\lambda$ (see Definition~\ref{def:surplus_split}). Building on this, \citet{chen2025evidence} developed a framework to estimate the surplus split factor from real-world mixed traffic and found that cars and trucks allocate the surplus with cars receiving 81\% and trucks 19\%. However, it is unclear how this manifests with AVs. We aim to estimate the surplus split in mixed autonomy traffic by solving for the optimal surplus split factor, $\lambda^*$.

We define a loss function aiming to best align the game-predicted speeds with the DRL simulated speeds as follows,
\begin{equation}\label{eq:lambda_optm_loss}
      L(\rho_1, \rho_2, \lambda; \lambda \in \left[0,1\right]) = \sum_{\rho_2=1}^{\rho_2^{max}} \sum_{\rho_1=1}^{\rho_1^{max}} 
 \left[ \sum_{i=1}^2 w_i \lvert \Bar{u}_i(\rho_1, \rho_2) - \hat{u}_i(\rho_1, \rho_2;\lambda)\rvert \right]^2
\end{equation}
Here, for more convenient calculations, we discretize the class density values into integers ranging from 1 to their respective maximum density values, $\rho_1^{max}$ and $\rho_2^{max}$. The term $w_i$ is the weighting factor that determines the relative importance between the errors of class 1 and class 2 speeds, and $\hat{u}_i(\rho_1, \rho_2;\lambda)$ is the speed predicted by the game theoretical model in (\ref{eq:payoff}). Additionally, $\Bar{u}_i(\rho_1, \rho_2)$ represents the average speed for class $i$ for the density pair $(\rho_1, \rho_2)$, calculated from the DRL evaluation data. Specifically, let $S$ be the set of all speed values for the density pair $(\rho_1, \rho_2)$, then the average speed is calculated as:
\begin{equation}\label{eq:u_i_bar}
    \Bar{u}_i(\rho_1, \rho_2) = \frac{1}{\lvert S \rvert} \sum_{k=1}^{\lvert S \rvert} u_k^S
\end{equation}
where $\lvert S \rvert$ is the size of set $S$, i.e., the number of speed values for density pair $(\rho_1, \rho_2)$, and $u_k^S$ is the $k$th speed value in $S$.

With (\ref{eq:lambda_optm_loss}), the goal is to find the optimal $\lambda$ that minimizes the loss,
\begin{equation}\label{eq:lambda_optm}
    \lambda^* =\arg \min_{\lambda} L(\rho_1, \rho_2, \lambda; \lambda \in \left[0,1\right])
\end{equation}

The estimation yields $\lambda^* = 0.6484$, with the mean absolute errors (MAE) for class 1, class 2, and the weighted average being 3.50, 2.70, and 3.10, respectively. This indicates that HVs receive 64.84\% of the cooperation surplus, while AVs receive 35.16\%. This suggests that in the DRL-attained Pareto-efficient NE, HVs benefit more than AVs. This could be attributed to the other benefits that AV passengers already gain from the autonomous driving technology, such as increased productivity or leisure time, reduced driving stress, and optimized fuel efficiency~\citep{rahman2023impacts}, making them less inclined to seek further road-based benefits.

\begin{figure}[ht] 
    \centering
    \includegraphics[width=.6\linewidth]{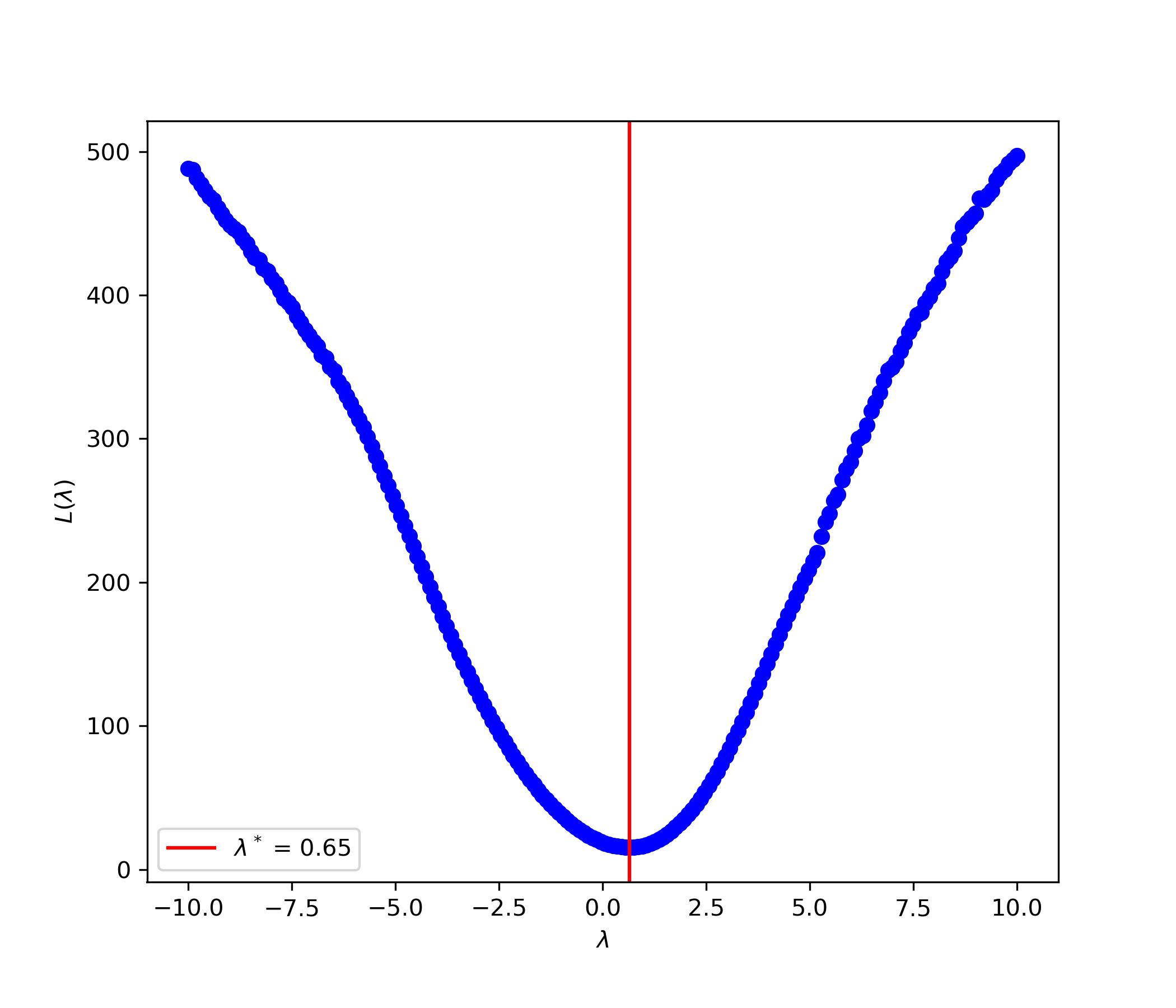}
    \caption{Optimal surplus split factor.}
    \label{fig:optimal_lambda}
\end{figure}

Moreover, Figure~\ref{fig:optimal_lambda} depicts the loss under different $\lambda$ values. Generally, the loss function appears symmetric around $\lambda^*$. This symmetry indicates that equilibrium can shift to both left and right with equal effort, suggesting that HVs and AVs have similar adaptability in receiving the cooperation surplus. 


The settings and solutions of the above parameters result in a Pareto-efficient NE from the bargaining game model that aligns with the DRL-attained NE. This benchmark model will be used for the subsequent analysis.

\subsubsection{First-order comparison}\label{sec:compare_game_first}

In the first-order comparisons of the Pareto-efficient NE between the DRL model and the benchmark game theoretical model, we aim to directly compare heatmaps for various traffic variables without performing mathematical calculations.

\begin{figure}[ht] 
    \centering
    \includegraphics[width=\textwidth]{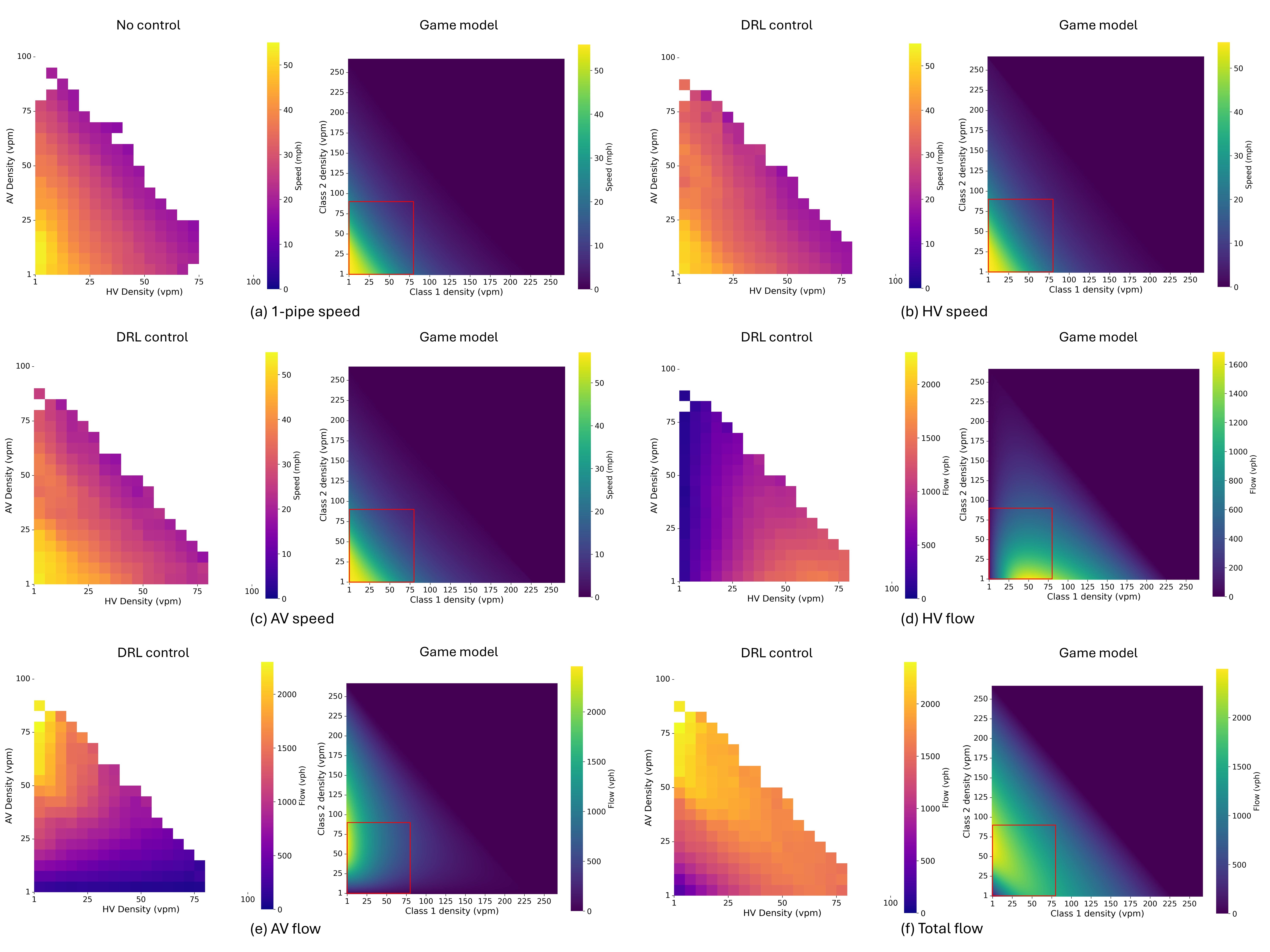}
    \caption{First-order comparisons of macroscopic traffic characteristics achieved by the DRL model and the Pareto-efficient equilibrium (collective rationality) predicted by the game theoretical model.}
    \label{fig:first_order}
\end{figure}

In Figure~\ref{fig:first_order}, panel (a) compares the 1-pipe equilibrium speed from the fully mixed (no control) scenario with the one solved from (\ref{eq:1pipe_speed}). Panels (b) and (c) compare class-specific speeds, while panels (d) and (e) compare class-specific traffic flow. The total flow of the mixed traffic is compared in panel (f). In the left figure for each panel, the speed value for each grid is calculated using (\ref{eq:u_i_bar}), and the flow value is computed by multiplying the speed value by the traffic density. Overall, the macroscopic traffic patterns observed in the simulation closely approximate the predictions from the benchmark model. This alignment indicates that, on one hand, CR is attainable through DRL, and on the other hand, this attainability is not coincidental but is grounded in a theoretical benchmark.

\subsubsection{Second-order comparison}\label{sec:compare_game_second}

The second-order comparison involves simple mathematical subtractions between different traffic variables to reveal their relative relationships. 

\begin{figure}[ht] 
    \centering
    \includegraphics[width=\textwidth]{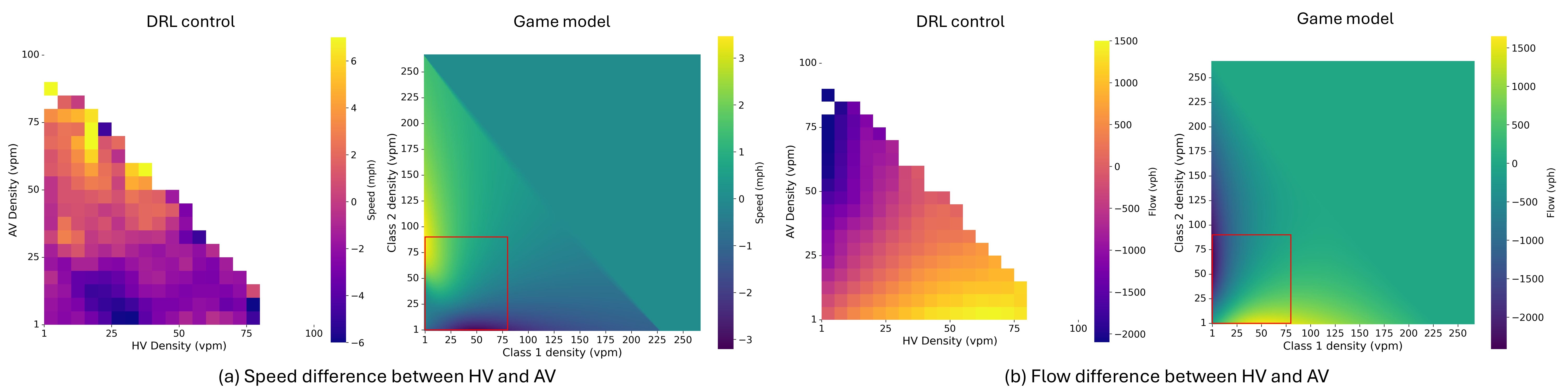}
    \caption{Second-order comparisons of speed and flow differences between HVs and AVs.}
    \label{fig:second_order_HV_AV}
\end{figure}

In Figure~\ref{fig:second_order_HV_AV}, we present the speed and flow differences between HVs and AVs, comparing these with the benchmark model's predictions. Overall, both speed and flow exhibit similar patterns to the benchmark model. However, the DRL model demonstrates greater heterogeneity. We seek to explain this phenomenon. Revisiting Figure~\ref{fig:first_order}(b)-(c), we denote the speed from DRL simulation as $\hat{u}_i(\rho_1, \rho_2)$ and the speed from the benchmark model as $u_i(\rho_1, \rho_2)$. Then their relationship can be expressed as,
\begin{equation}
\left\{
  \begin{array}{ll}
    \hat{u}_1(\rho_1, \rho_2) = u_1(\rho_1, \rho_2) + \epsilon_1 \\
    \hat{u}_2(\rho_1, \rho_2) = u_2(\rho_1, \rho_2) + \epsilon_2
  \end{array}
\right.
\end{equation}
where $\epsilon_i$ is the error between the DRL simulation data and the game model's predictions. Denoting the variance of errors $\epsilon_1$ and $\epsilon_2$ as $\sigma_1$ and $\sigma_2$, respectively, the speed difference between HV and AV is computed as,
\begin{equation}
    \Delta = \hat{u}_1 - \hat{u}_2
\end{equation}
For simplification, assuming $\hat{u}_1$ and $\hat{u}_2$ are independent, then the randomness for $\Delta$ becomes $\epsilon_1 + \epsilon_2$, with the variance of randomness being $\sigma_1 + \sigma_2$. This accumulation of randomness in $\Delta$ leads to discrepancies in Figure~\ref{fig:second_order_HV_AV}(a) compared to the benchmark model. 
In contrast, as illustrated in Figure~\ref{fig:second_order_HV_AV}(b), the flow difference is less affected by such randomness. The effect of speed variations is smoothed out over the density of vehicles, a deterministic variable that acts as a stabilizing factor, leading to a reduction of the randomness observed in the magnitude of flow. 

\begin{figure}[ht] 
    \centering
    \includegraphics[width=\textwidth]{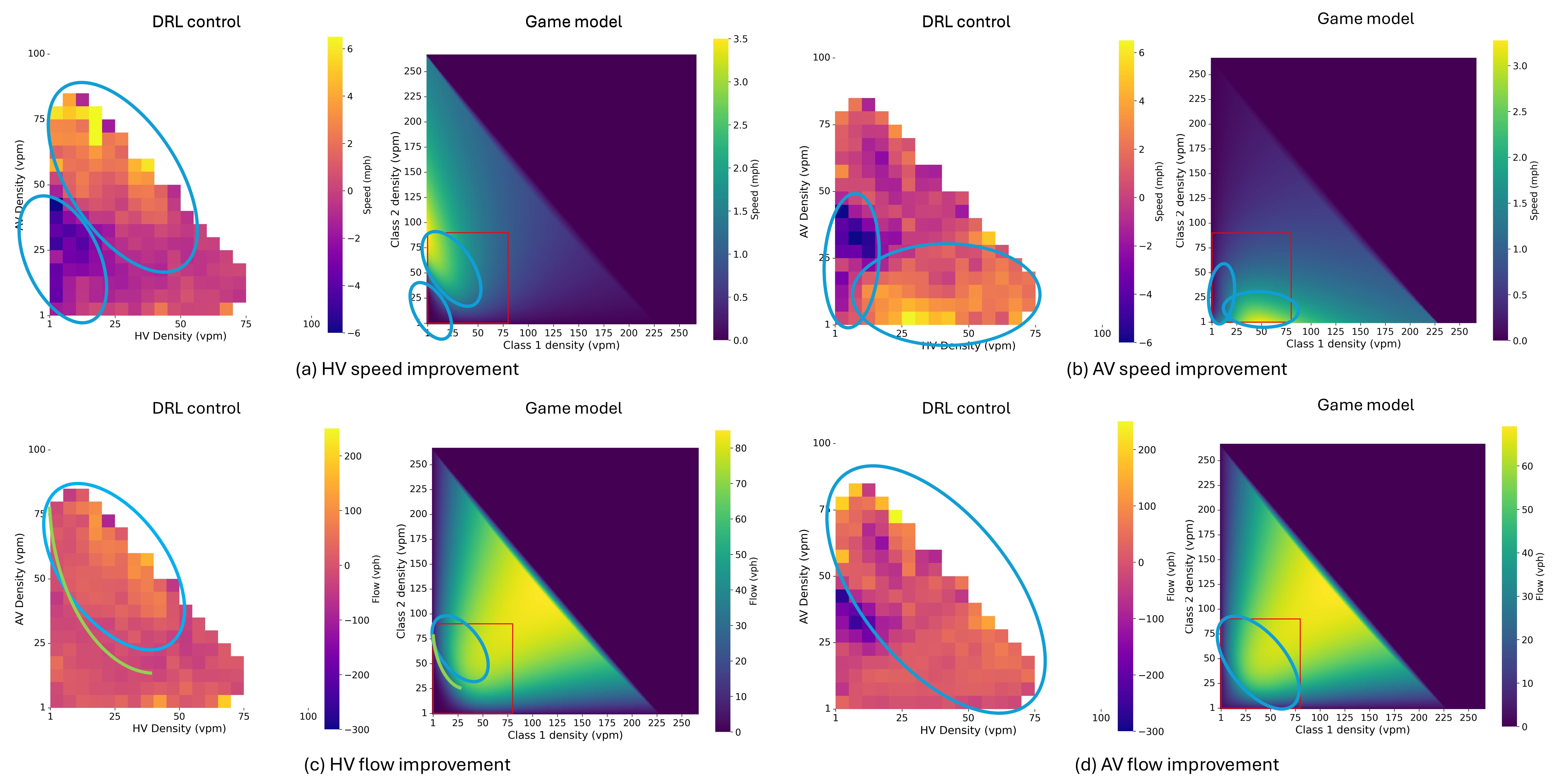}
    \caption{Second-order comparisons of speed and flow improvements between HVs and AVs.}
    \label{fig:second_order_s_nos}
\end{figure}

In Figure~\ref{fig:second_order_s_nos}, we compare the improvements in speed and flow relative to the fully mixed scenario (i.e., no CR). We acknowledge the observation of some differences, due to the inherent heterogeneity in simulation. Nonetheless, notable similarities are also present, highlighted by blue circles indicating regions of similarity and the green curve marking the boundary of positive and non-positive value transitions. For instance, in Figure~\ref{fig:second_order_s_nos}(a), both the DRL-attained NE and the theoretical prediction show positive speed improvement at higher AV densities. Similarly, in Figure~\ref{fig:second_order_s_nos}(c), the boundaries where values transition from positive to non-positive appear similar. Overall, these similarities suggest that the possibility of CR's emergence from the DRL model.
Furthermore, the DRL model offers richer outcomes compared to the theoretical model. For example, the DRL model captures negative improvements on speed and flow due to uncertainties present in both DRL-controlled and no-control simulation environments. While the theoretical model, with its simplified assumptions, fails to account for these variations.

\subsubsection{Fundamental diagram comparison}\label{sec:compare_game_fd}

\begin{figure}[ht] 
    \centering
    \begin{subfigure}{.49\textwidth}
        \centering
        \includegraphics[width=1\linewidth]{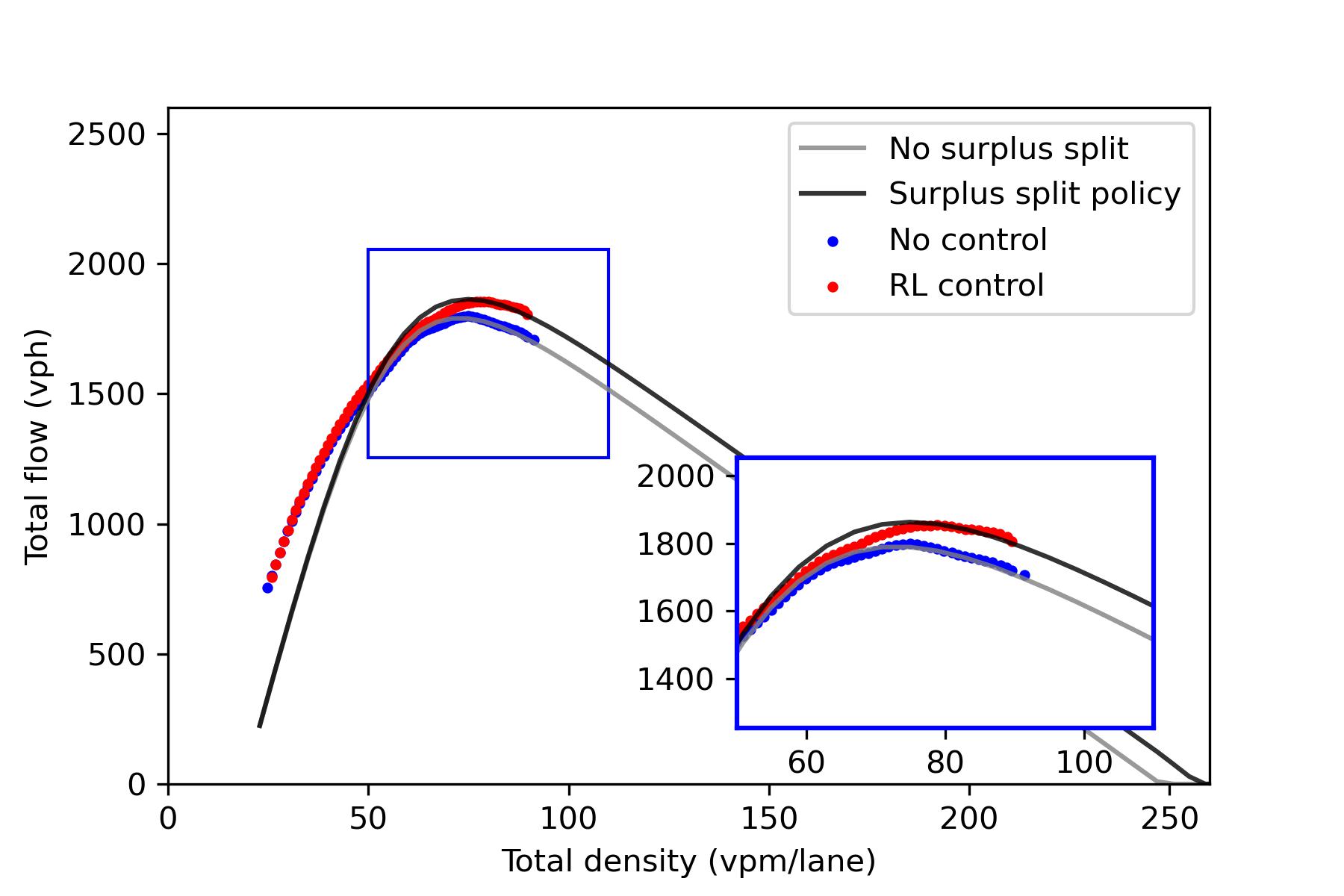}
        \caption{25\% AV}
    \end{subfigure}
     \begin{subfigure}{.49\textwidth}
        \centering
        \includegraphics[width=1\linewidth]{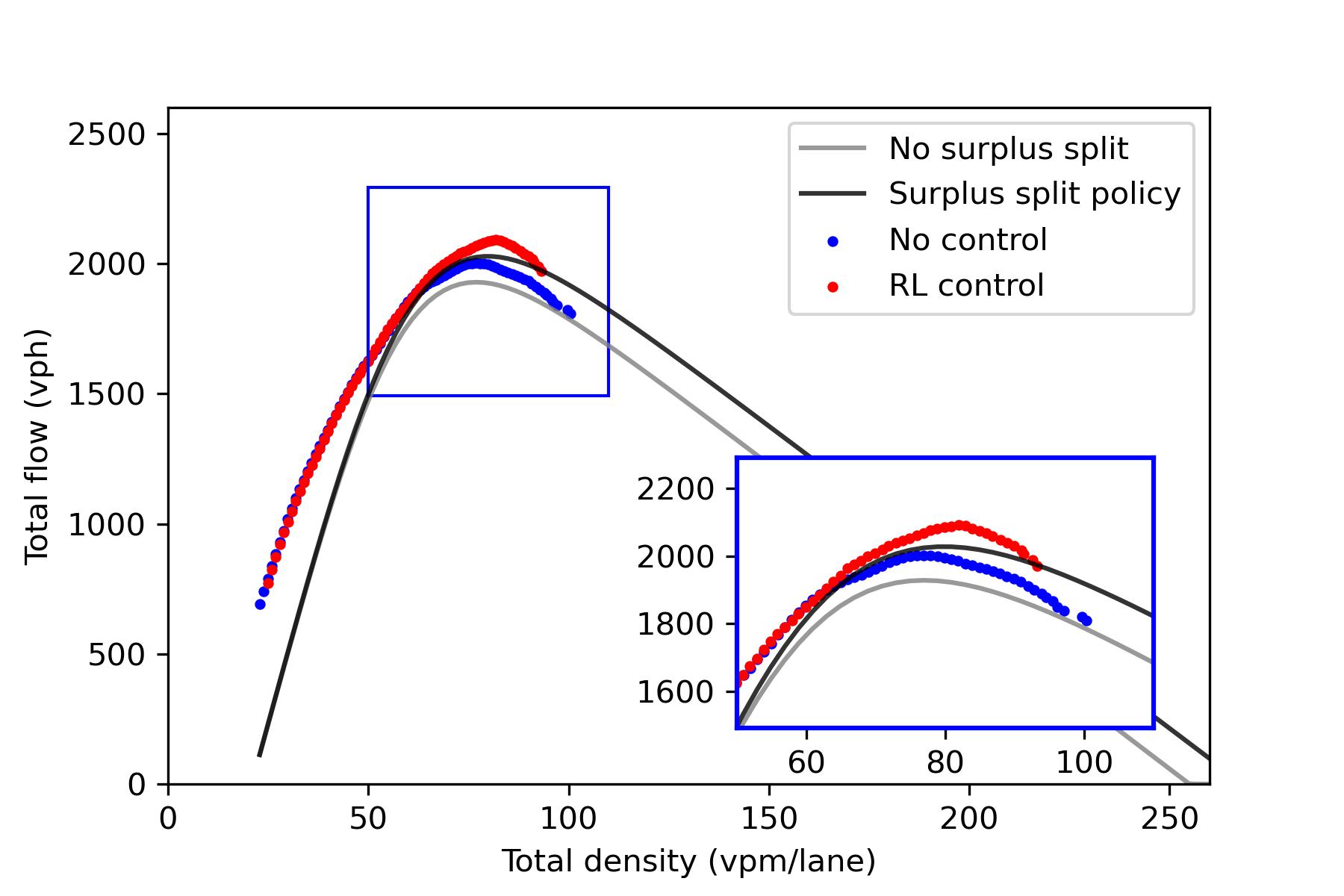}
        \caption{50\% AV}
    \end{subfigure}
     \begin{subfigure}{.49\textwidth}
        \centering
        \includegraphics[width=1\linewidth]{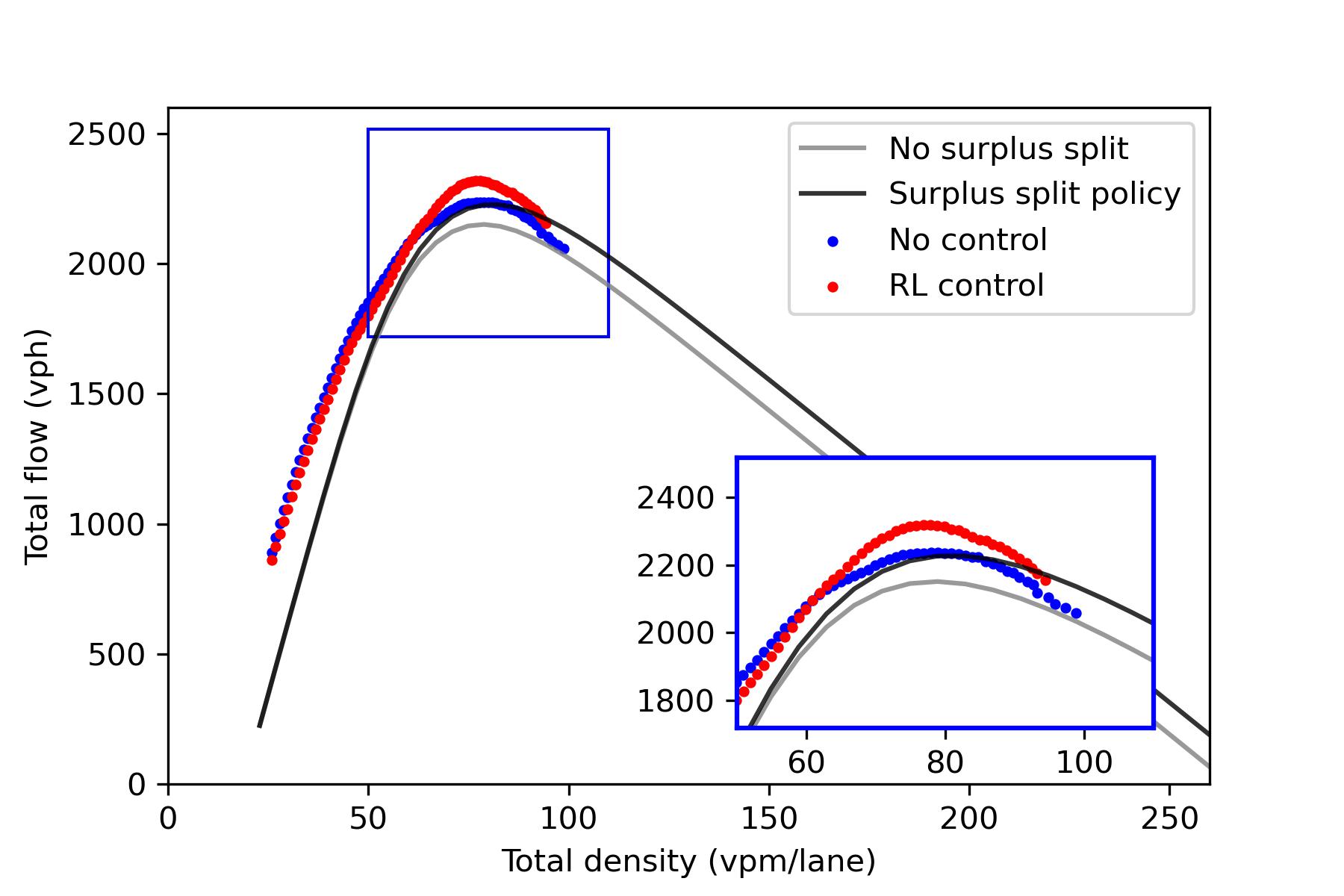}
        \caption{75\% AV}
    \end{subfigure}
    \caption{Comparison of fundamental diagrams between the DRL model and game theoretical model under different AV penetration rates.}
    \label{fig:q-k}
\end{figure}

The comparisons of fundamental diagrams are presented in Figure~\ref{fig:q-k}, where both flow and density represent the combined totals for the two vehicle classes. As indicated by the black and gray lines, theoretically, the existence of CR improves traffic flow due to the fully utilized cooperation surplus compared with the fully mixed regime~\citep{li2022equilibrium}. To verify this for the CR attained by DRL, we plot the flow-density relations using the simulation data, as represented by red (DRL control) and blue (no control) scatter points in Figure~\ref{fig:q-k}. The maximum flow under DRL control is higher than no DRL control, especially near critical densities, aligning with theoretical predictions. The improvements in maximum flow are 90 vph/lane for 25\% AVs, 130 vph/lane for 50\% AVs, and 98 vph/lane for 75\% AVs. Besides, the maximum flow tends to be higher with increased AV penetration rates for both the theoretical and DRL models. 
These alignments further support our hypothesis on the attainability of CR through DRL.



\section{Mechanisms to attain collective rationality}\label{sec:mechanism}

\subsection{Hypothesized mechanism}

\begin{figure}[!ht] 
    \centering
    \includegraphics[width=\textwidth]{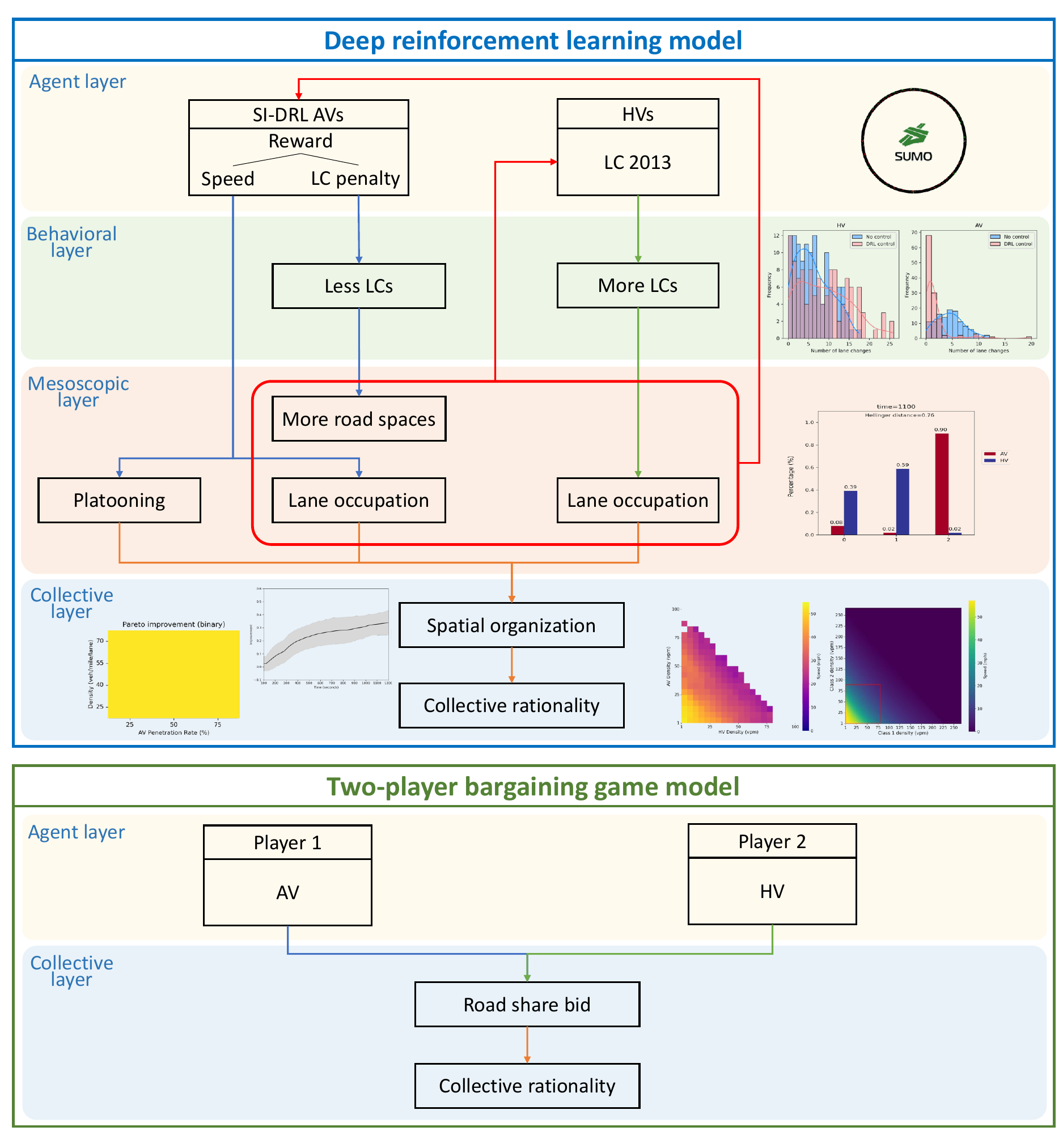}
    \caption{Hypothesized mechanisms for attaining collective rationality through deep reinforcement learning and comparisons with the game theoretical model. The term ``lane change'' is referred to LC in the figure. } 
    \label{fig:RL_CR_causal}
\end{figure}

Drawing on the evidence on the emergence of CR through DRL, it is natural to ask what factors contribute to this attainment in the microscopic and dynamic environment. We hypothesize a mechanism to explain the formation of CR, which is illustrated in Figure~\ref{fig:RL_CR_causal}. In contrast to the game theoretical model, which is a “one-shot” bargaining focused on the macroscopic scale and all possible static equilibria, the DRL method elucidates the dynamic process of reaching CR. The core hypothesis posits that DRL, even with a simple reward design, can achieve better spatial organization through lane-changing maneuvers. This enhanced spatial organization translates to Pareto-efficient NE, indicating the attainment of CR. 

As shown in Figure~\ref{fig:RL_CR_causal}, the DRL simulation environment captures the dynamic process of interactions between AVs and HVs. For SI-DRL AVs, the lane change penalty reward leads to fewer lane changes among AVs. Since lane changes impact both the current and target lanes~\citep{jin2010kinematic}, this reduction by AVs facilitates easier lane changes for other vehicles. Meanwhile, the speed reward encourages SI-DRL AVs to form platoons and occupy preferred lanes. HVs, which make lane-change decisions based on SUMO's built-in model, respond to AVs' lane occupation by making more lane changes to optimize their lane positions. These changes in lane occupation subsequently influence AVs' decision-making, creating a feedback loop that reflects the dynamic interplay between AVs and HVs. Depending on whether the feedback loop is self-reinforcing (influenced by decision-making of both HVs and AVs), the system eventually converges to either a CR or non-CR state at the collective level.


The subsequent sections will validate the hypothesized mechanism based on simulation evidence. This includes analyzing lane-changing behaviors at the behavioral layer (Section~\ref{sec:LC}), examining platooning and lane occupations at the mesoscopic layer (Section~\ref{sec:meso_layer}), and introducing a spatial organization metric to assess macroscopic spatial organization in the collective layer (Section~\ref{sec:spatial}). Additionally, we will further elucidate the relationship between the proposed spatial organization metric and the game theoretical model in the collective layer (Section~\ref{sec:spatial_surplus}).


\subsection{Behavioral analysis}\label{sec:LC} 

\begin{figure}[!ht] 
    \centering
    \includegraphics[width=\textwidth]{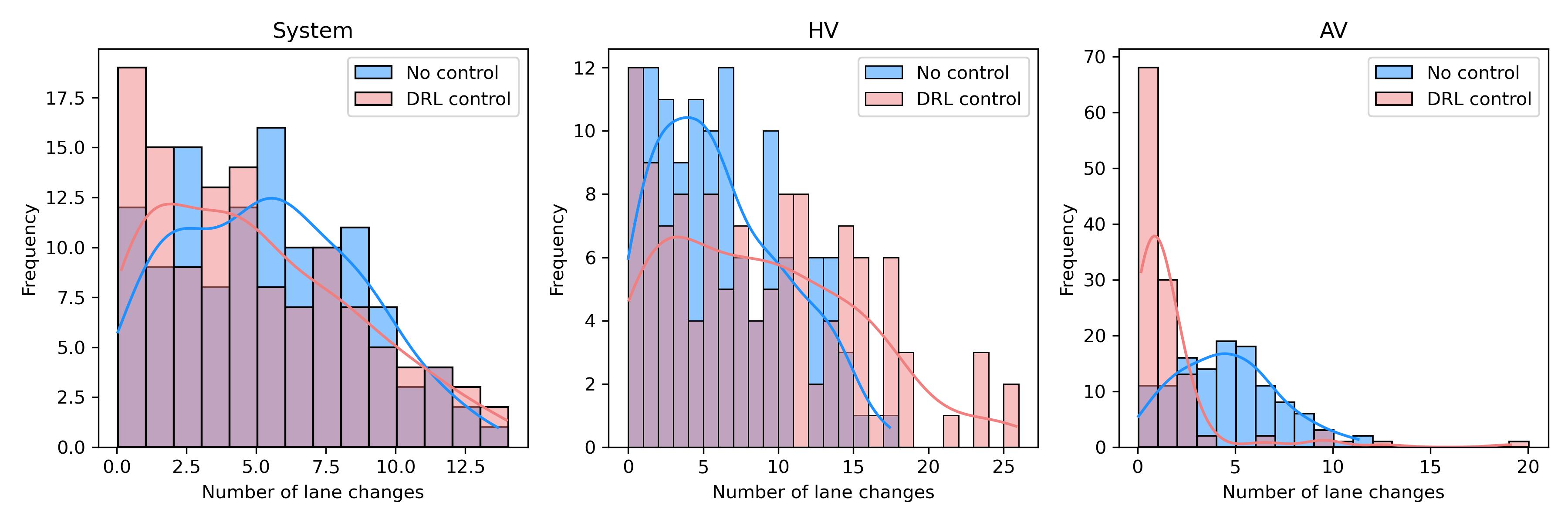}
    \caption{Comparisons of lane change frequency distributions between DRL control and no control.} 
    \label{fig:LC_compare}
\end{figure}

As postulated in Figure~\ref{fig:RL_CR_causal}, the SI-DRL AVs engage in fewer lane change maneuvers, leading to HVs making more lane changes during their interactions. To assess these lane change behaviors, we measure the lane change frequency for each class and compare the distributions with the no-control scenario. The lane change frequency is calculated as the average number of lane changes per vehicle over each evaluation period, excluding the vehicle loading time. Figure~\ref{fig:LC_compare} illustrates lane change frequency distributions for the system, HVs, and AVs. The results indicate significant differences between DRL control and no-control scenarios for both AVs and HVs. Specifically, in the DRL-controlled environment, SI-DRL agents perform fewer lane changes due to their optimized decision models, while HVs show more lane changes to adapt to AV impacts and find optimal lanes. These interactions suggest the potential for different spatial distributions compared to the no-control scenario.

\subsection{Mesoscopic analysis}\label{sec:meso_layer}



\begin{figure}[!ht]
    \centering
    \includegraphics[width=\textwidth]{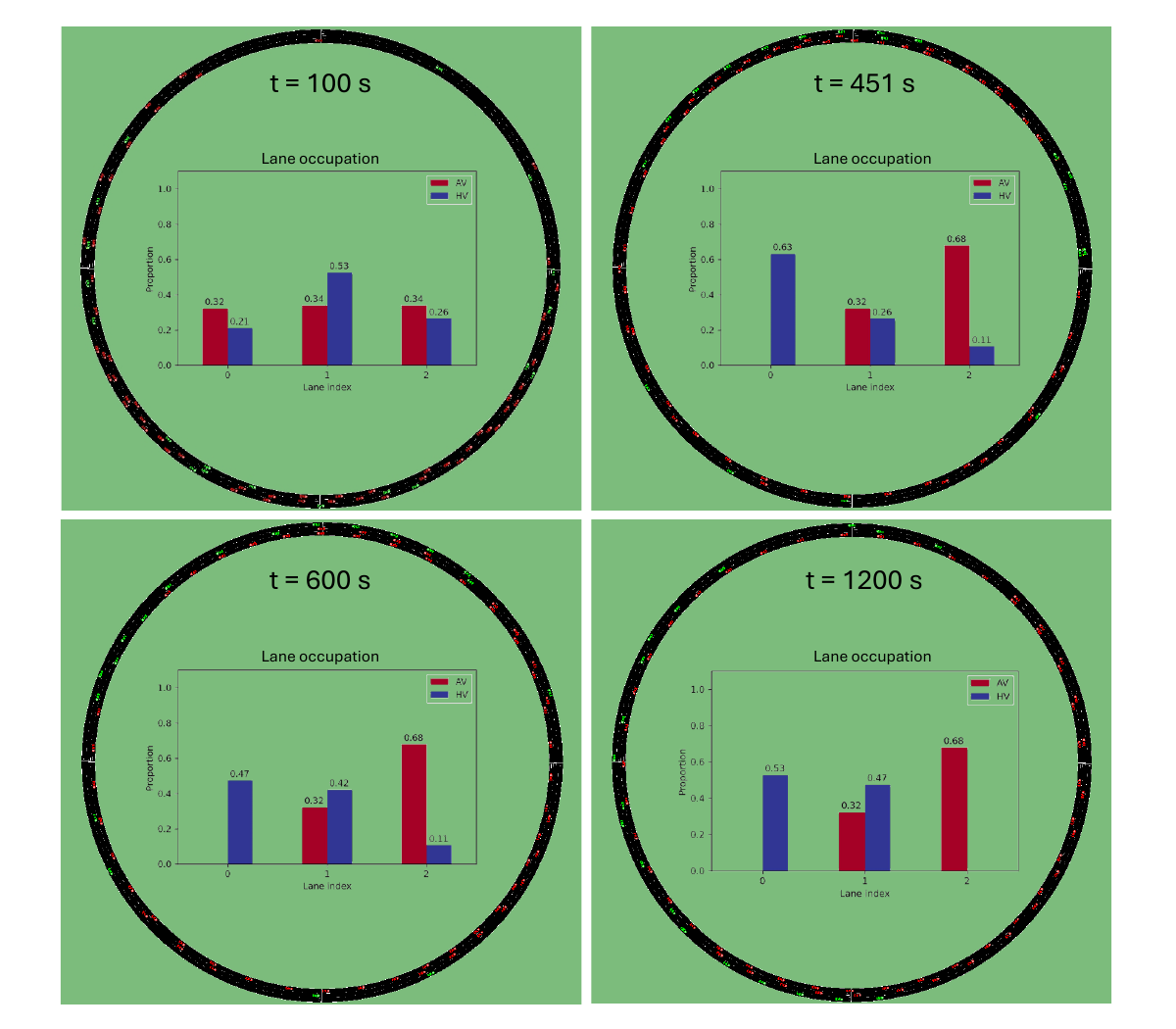}
    \caption{Screenshot of evaluation scenario for 40 vpm/lane and 75\% AVs in SUMO. Vehicles in red and green represent AVs and HVs, respectively.}
    \label{fig:sumo}
\end{figure}

To analyze the AV-HV interactions at the mesoscopic level, we capture screenshots from the DRL model evaluation process in SUMO and examine the lane occupation and platooning behaviors of vehicles. Figure~\ref{fig:sumo} illustrates a typical scenario with a traffic density of 40 vpm/lane and 75\% AVs. 
At timestamp $t = 100$, vehicles have just been loaded onto the road, and the system is in a fully mixed state, with both vehicle classes approximately evenly distributed across lanes and AVs not yet forming large platoons.
At timestamp $t = 451$, equilibrium is reached and maintained. Different lane occupations between the two classes start to emerge. AVs begin to occupy their preferred lanes, specifically lane 1 (middle lane) and lane 2 (leftmost lane). In response to the lane occupation by AVs, more HVs shift to the rightmost lane. Additionally, AVs start forming platoons, but the platoon sizes are still small and some AVs remain outside of platoons.
At timestamp $t = 600$, AVs maintain their lane occupation, while HVs continue to explore different lane positions. Larger platoons are also forming among AVs.
Finally, at $t = 1200$, AVs and HVs predominantly occupy separate lanes while sharing the middle lane. In the middle lane, AVs are platooned together, traveling separately from HVs.

\subsection{Spatial organization of mixed traffic}\label{sec:spatial}

Building on our previous work~\citep{li2022equilibrium}, which showed that CR emerges in traffic regimes where the two classes of vehicles are spatially separated, we postulate that when CR is present in this study, AVs and HVs may exhibit different spatial organization at the collective level, as illustrated in Figure~\ref{fig:RL_CR_causal}. This section seeks to validate this hypothesis by examining the spatial organization of mixed traffic through simulation evidence. To this end, we propose a spatial organization metric that evaluates the evolution of spatial arrangements, quantifying both intra-class platooning and inter-class segregation behaviors across longitudinal and lateral dimensions. 

\subsubsection{The proposed spatial organization metric}\label{sec:spatial_metric}

In mixed traffic, the spatial distribution between vehicle classes impact the overall system performance~\citep{ma2021analysis, yao2022fundamental}. Our previous work~\citep{li2022equilibrium} also analyzed different configurations of mixed traffic in the analytical modeling of CR, finding that CR is present in a regime where the two vehicle classes are spatially separated from each other. Therefore, to understand the underlying mechanisms by which DRL achieves CR, we propose a spatial organization metric to evaluate the evolution of spatial arrangements in mixed traffic as it progresses towards CR. This metric quantifies both intra-class platooning and inter-class segregation behaviors, encompassing both longitudinal and lateral dimensions.

\bigskip
(i) Intra-class metric
\bigskip

As shown in Figure~\ref{fig:conceptual}, the benefits of CR arise from more AVs platooning together, where AVs' smaller desired headways result in road space savings. Consequently, the intra-class metric, which focuses specifically on AVs, aims to quantify their platooning behaviors.

Platooning refers to a fleet of vehicles traveling with a small desired inter-vehicle spacing with the help of advanced automated driving systems. Typical definitions for such inter-vehicle spacing include the constant spacing (CS) policy and the constant time headway (CTH) policy. The CS policy sets a fixed inter-vehicle distance, whereas the CTH policy sets a fixed time headway and adjusts the inter-vehicle distance based on vehicle speed. We adopt the CTH policy as it better addresses safety concerns by varying the spacing with travel speed. Under the CTH policy, the maximum spacing considered as part of a platoon is defined as,
\begin{equation}
    s_p^* = u h_{AA}^* + d_0
\end{equation}
where $s_p^*$ is the equilibrium inter-vehicle spacing, $u$ is the vehicle travel speed, $h_{AA}^*$ is the desired equilibrium time headway between AVs, and $d_0$ is the standstill gap. In this study, we adopt $h_{AA}^*=1 \ sec$ and $d_0=1 \ m$.

Vehicles in different platoon positions experience varying benefits, with platoon members generally receiving more advantages, while the platoon leader benefits less~\citep{tsugawa2016review, shladover2015cooperative}. To focus primarily on the mechanisms of CR, we simplify the platooning problem by assuming that only platoon members gain benefits. Accordingly, the intra-class metric quantifies the percentage of AVs benefiting from platooning and consists of two components: one measures the percentage of AVs traveling within platoons, denoted as $P_e \in [0,1]$, and the other measures the percentage of AVs acting as platoon leaders, denoted as $P_s \in [0, 1/2]$. The objective is to encourage the formation of platoons while penalizing the creation of numerous small platoons. The benefit of platooning, denoted as $B$, is given by,
\begin{equation}\label{eq:B_t}
    B = \frac{N_{AV}^{p} - N_{p}}{N_{AV}} = Pe - Ps
\end{equation}
where $N_{AV}$ denotes the total number of AVs, $N_{AV}^P$ denotes the number of AVs in platoons, and $N_p$ represents the number of platoons, which is equivalent to the number of platoon leaders. 

\bigskip
(ii) Inter-class metric
\bigskip

The inter-class metric assesses the spatial distribution between HVs and AVs. In measuring the spatial distributions of mixed traffic, studies often employ statistical methods such as probability theory to quantify the likelihood of platooning~\citep{chang2020analysis, ma2021analysis}. However, these metrics typically focus solely on the longitudinal dimension and tend to overlook the lateral distribution (i.e., lane distribution) of mixed traffic agents.

To quantify the spatial distribution between the two classes both longitudinally and laterally, we utilize a statistical metric known as the Hellinger distance~\citep{hellinger1909neue}. This metric measures the similarity between two probability distributions, either continuous or discrete, without requiring specific assumptions about the distributions. Additionally, it can handle multivariate distributions~\citep{tamura1986minimum, cutler1996minimum}, making it suitable for comparing two sets of bivariate distributions.

To begin with, we define a one-dimensional Hellinger distance to measure the differences in lane distributions between vehicle classes. Let $p_i$ and $q_i$ represent the percentage of AVs and HVs in the $i$th lane relative to the total number of AVs and HVs across all lanes, respectively. And denote $P=\{p_i \mid i=1, \dots,N\}$ and $Q=\{q_i \mid i=1, \dots, N\}$ as two discrete probability distributions for AVs and HVs respectively, and it follows that $\sum_{i=1}^N p_i=1$ and $\sum_{i=1}^N q_i=1$, where $N$ is the total number of lanes.
Then, the Hellinger distance is given by:
\begin{equation}\label{eq:H_1D}
    H(P, Q) = \frac{1}{\sqrt{2}} \sqrt{\sum_{i=1}^N (\sqrt{p_i} - \sqrt{q_i})^2}
\end{equation}
This distance measures the extent to which AVs and HVs are laterally separated from each other. It satisfies $H(P, Q) \in [0, 1]$, with 0 represents the minimum distance (fully mixed) and 1 represents the maximum distance (fully separate). Examples of these extreme cases are illustrated in Figure~\ref{fig:conceptual}. 

Such lateral organization of mixed traffic was also discussed in \citet{li2022equilibrium} in terms of fully mixed and fully separate regimes, which conceptually aligns with what the one-dimensional Hellinger distance captures. However, \citet{li2022equilibrium} overlooks longitudinal spatial organization due to its simplified assumptions, which should not be neglected in the more realistic settings of this study. To address this, we extend the Hellinger distance to a two-dimensional case.


We partition the multi-lane ring road into $M \times N$ cells, where $M$ represents the number of cells in each lane longitudinally. Each cell contains a certain number of AVs and HVs. The proportion of AVs in cell $(i, j)$ over the total number of AVs is denoted as $p_{ij}$; similarly of $q_{ij}$ for HVs. Then the probability distributions for AVs and HVs are respectively represented as $P=\{p_{ij} \mid i=1, \dots,N; j=1,\dots,M \}$ and $Q=\{q_{ij} \mid i=1, \dots,N; j=1,\dots,M\}$. The two-dimensional Hellinger distance is therefore written as,
\begin{equation}\label{eq:H_2D}
    H(P, Q) = \frac{1}{\sqrt{2}} \sqrt{\sum_{i=1}^N \sum_{j=1}^M \left(\sqrt{p_{ij}} - \sqrt{q_{ij}} \right)^2}
\end{equation}
Here, the relation follows $\sum_i \sum_j p_{ij}=1$ and $\sum_i \sum_j q_{ij}=1$. As a special case of this definition, by setting $M=1$, meaning each lane is considered a cell, (\ref{eq:H_2D}) reduces to the one-dimensional Hellinger distance as defined in (\ref{eq:H_1D}). For the studied ring road scenario, we set $N=3$ and $M=10$. 


\bigskip
(iii) Integrated spatial organization metric
\bigskip

By integrating the proposed metrics in (\ref{eq:B_t}) and (\ref{eq:H_2D}), we can assess the degree of spatial organization at any timestamp $t$. The spatial organization metric $M_t$ is expressed as,
\begin{equation}\label{eq:eval_metric}
    M_t = a_1 B_t + a_2 H_t
\end{equation}
where the coefficients $a_1$ and $a_2$ determine the emphasis on micro-scale (intra-class) versus system-level (inter-class) performance. In this study, we select $a_1 = 0.5$ and $a_2 = 1$.

\subsubsection{Spatial organization evaluation}\label{sec:spatial_eval}

\begin{figure}[ht] 
    \centering
    \begin{subfigure}{.49\textwidth}
        \centering
        \includegraphics[width=1\linewidth]{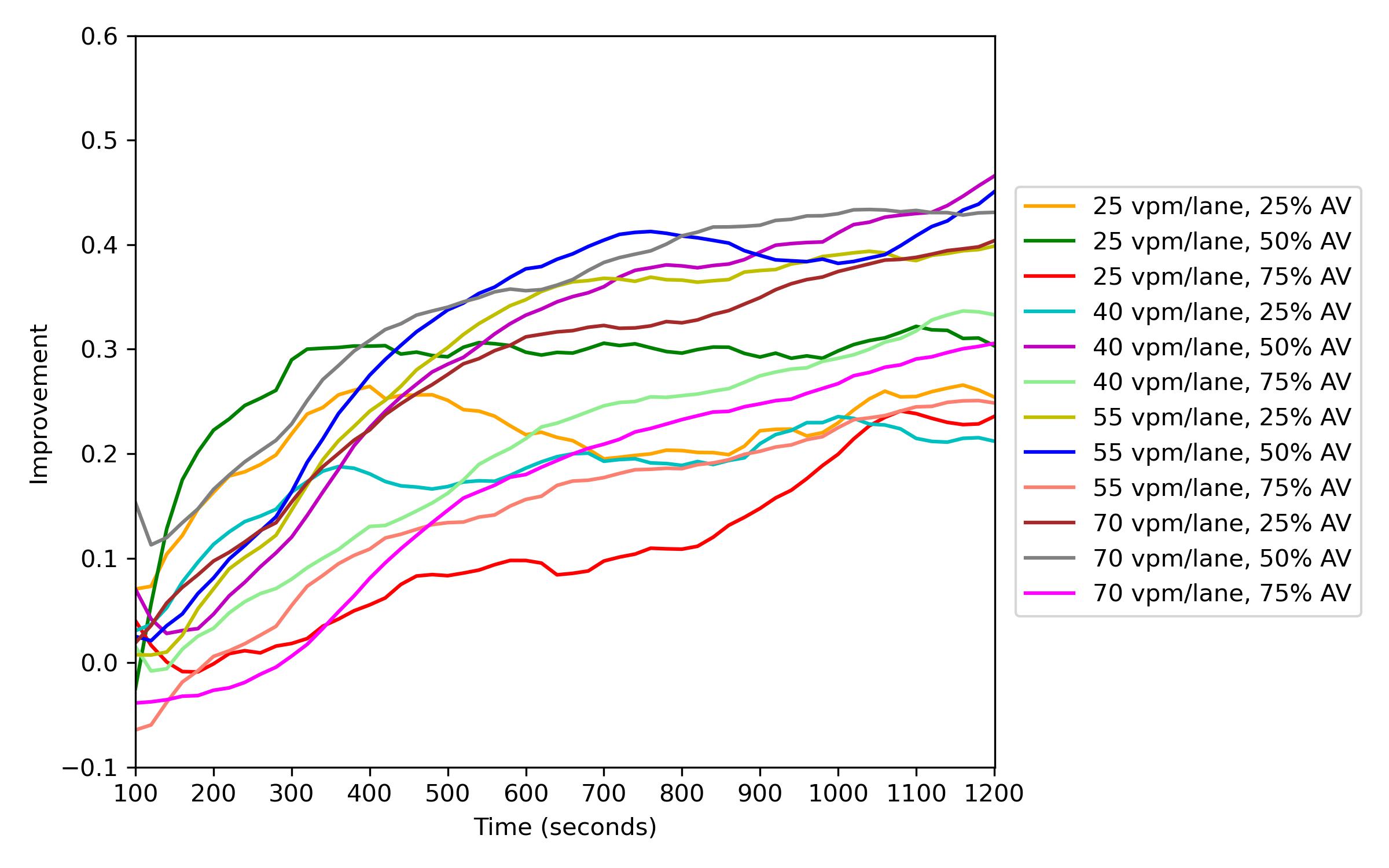}
        \caption{Improvement of spatial organization across traffic scenarios}
        \label{fig:metric_2D}
    \end{subfigure}
    \begin{subfigure}{.49\textwidth}
         \centering
        \includegraphics[width=1\linewidth]{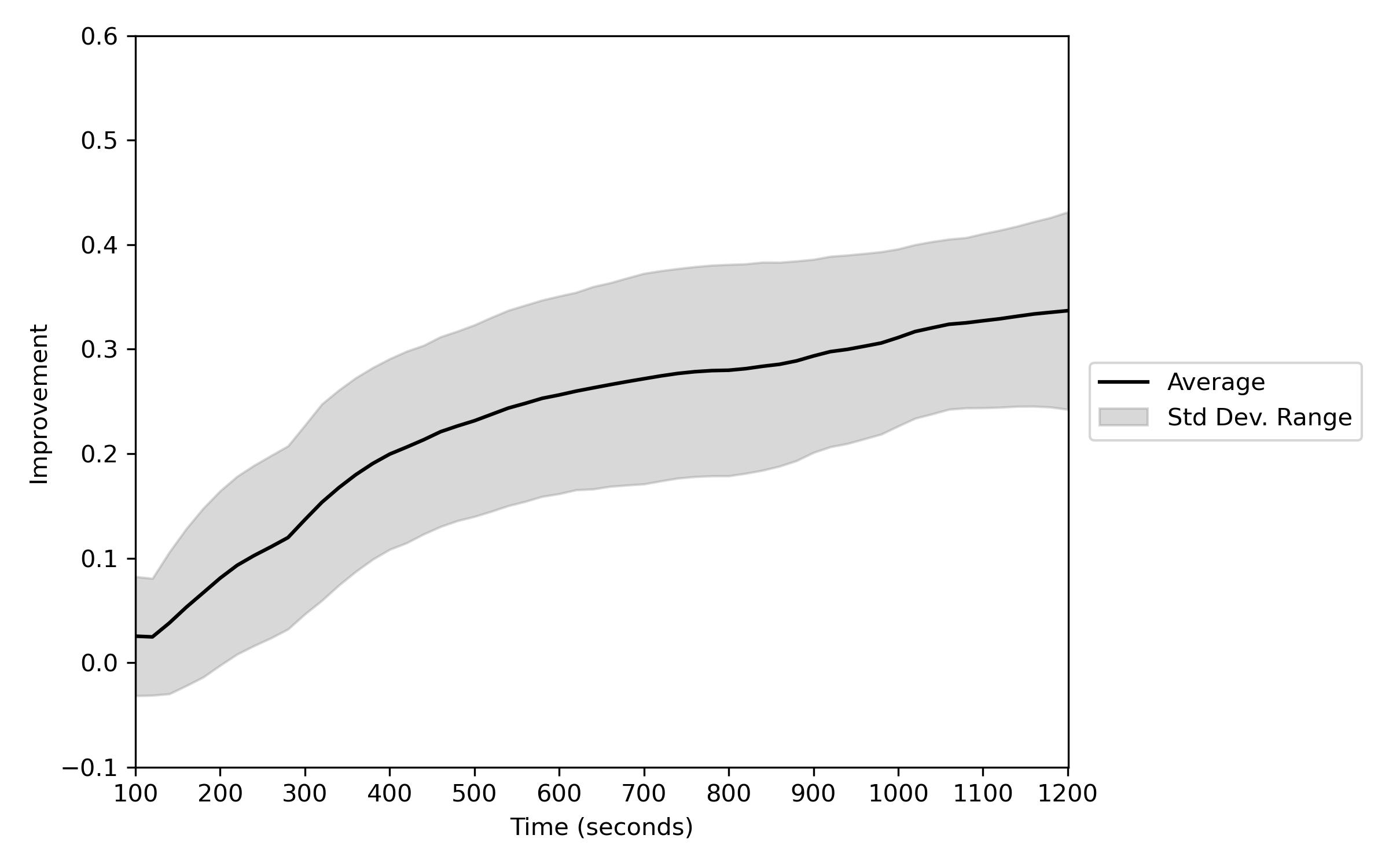}
        \caption{Mean and standard deviation across traffic scenarios}
        \label{fig:metric_over_time_agg}
    \end{subfigure}
    \caption{Improvement of the spatial organization metric (\ref{eq:eval_metric}) under DRL control compared to no control. The first 100 seconds are excluded as they represent the vehicle loading time.} 
    \label{fig:metric_over_time}
\end{figure}

The evolution of spatial organization between HVs and AVs is depicted in Figure~\ref{fig:metric_over_time}. The y-axis represents the difference in metric (\ref{eq:eval_metric}) between the DRL-controlled scenario and the no-control baseline (i.e., fully mixed). The metric generally remains positive and increases over time, indicating a tendency for mixed traffic to become more spatially organized in both longitudinal and lateral dimensions. An exception is observed at a density of 25 vpm/lane with 25\% AVs. This occurs because, under free-flow conditions with ample road space available, self-interested agents can benefit without the need for spatial organization. Despite this exception, the overall trend across all traffic conditions shows a tendency towards spatial organization, as indicated by Figure~\ref{fig:metric_over_time_agg}.

\subsection{Relations with the game theoretical model}\label{sec:spatial_surplus}

The improvement of the proposed spatial organization metric (\ref{eq:eval_metric}) indicates a more organized use of road space in the DRL-controlled scenario compared to fully mixed traffic. This concept is similar to the cooperation surplus (see Definition~\ref{def:surplus}), where players achieve a more organized traffic regime to be better off than in a fully mixed regime. Therefore, it is natural to explore if these two variables are related and how they correlate.

\begin{figure}[ht]
    \centering
    \includegraphics[width=.7\textwidth]{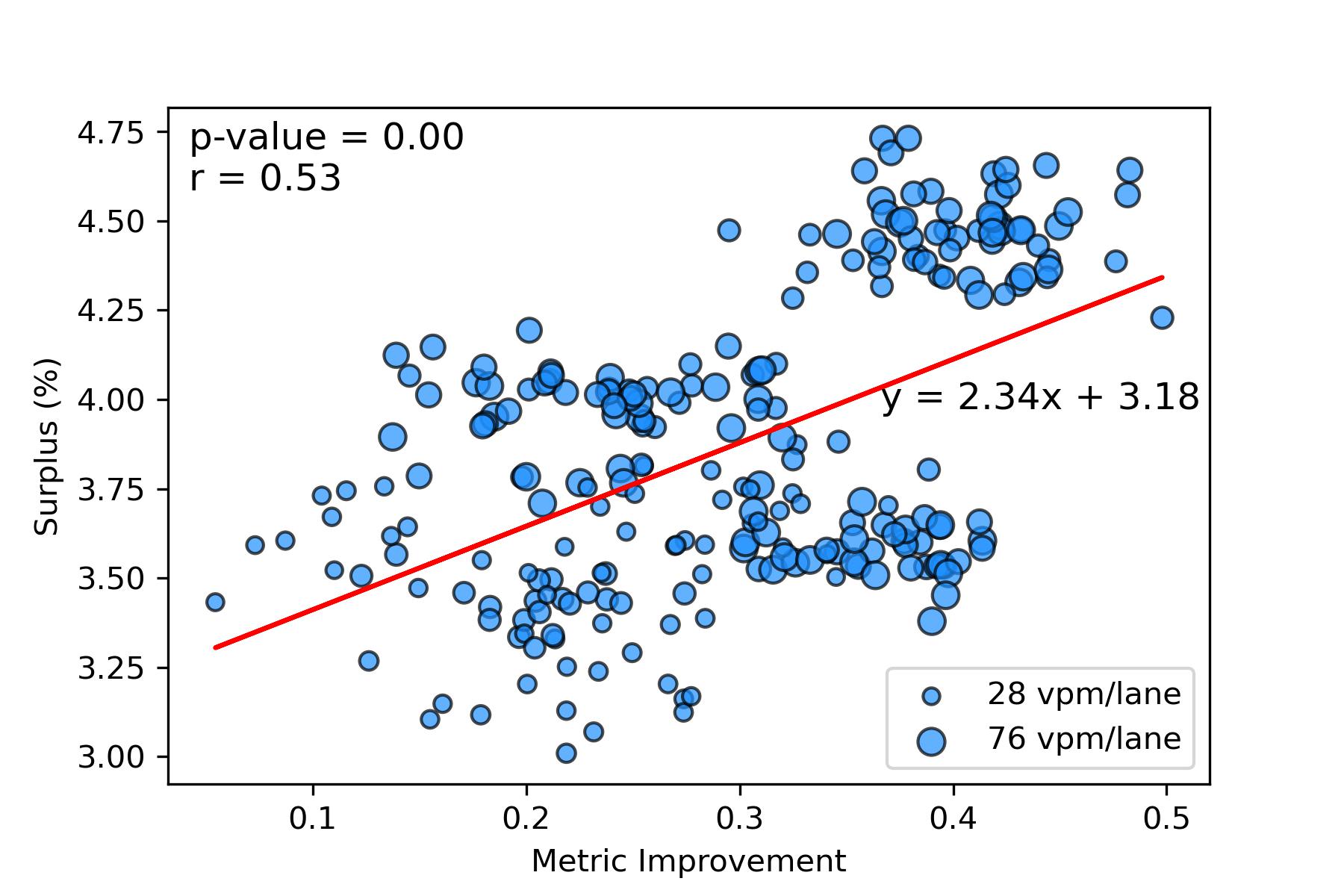}
    \caption{Relationship between cooperation surplus and improvement of the spatial organization metric.}
    \label{fig:surplus_metric}
\end{figure}

To connect these two variables, we align the traffic densities of each class from the DRL simulation output with the corresponding traffic densities from the game theoretical model.
The relationship between the predicted cooperation surplus and the improvement in spatial organization is illustrated in Figure~\ref{fig:surplus_metric}. A positive linear relationship is observed, where the cooperation surplus (in \%) increases by 2.34 units for every unit increase in spatial organization improvement, with a y-intercept of 3.18. The p-value of 0 indicates that this relationship is statistically significant, indicating a extremely low probability of observing this data by chance. Besides, the Pearson's correlation coefficient of 0.53 indicates a moderate positive correlation between these two variables. 

Furthermore, the size of the scatter points in Figure~\ref{fig:surplus_metric} represents total traffic densities. Smaller points are mainly clustered in the lower-left corner, while larger points are in the upper-right corner. This suggests that within the range of the experimented traffic densities, CR tends to be more evident at higher traffic densities.

Overall, this positive relationship implies the robustness of CR's emergence from both analytical models and simulation-based models. Moreover, it suggests that enhancing spatial organization is a viable approach to achieving CR in real-world mixed traffic scenarios.

\section{Conclusion}\label{sec:conclusion}


This study examines how self-interested AV behaviors can positively impact mixed traffic systems, through the lens of collective rationality (CR). Using DRL with a simple reward design, we demonstrate that CR can emerge in microscopic and dynamic mixed autonomy environment. Comparisons between the DRL-based results and our earlier game-theoretical predictions~\citep{li2022equilibrium} show the first-order alignment between the two, and richer second-order outcomes from DRL-driven learning process.
The analysis of fundamental diagrams shows a maximum capacity improvement of 130 vph/lane under 50\% AVs when collective rationality is attained. Moreover, the surplus split factor estimation reveals HVs gain 64.84\% of benefits from collective rationality, with AVs taking 35.16\%. 

We further hypothesized and validated an underlying mechanism for the emergence of CR in the physical driving environment. Microscopically, significant differences in lane-changing behaviors are observed for both AVs and HVs when collective rationality is attained. Macroscopically, the proposed spatial organization metric shows a stable increasing trend in both inter-class segregation and intra-class platooning behaviors. These results reveal that collective rationality emerges as mixed traffic agents perform lane-changing and car-following maneuvers in a way that results in a more organized traffic system. Overall, these findings suggest that enhancing spatial organization in real-world mixed traffic has the potential to benefit all road users.

We envision several extensions for future work. Firstly, advanced learning methods, such as federated learning, can be adopted to achieve collective rationality among self-interested driving agents while ensuring data privacy for AVs. This is achieved by allowing each AV to train its local model using local data while only sharing the model parameters to the central aggregator. Secondly, the lane-changing decisions for HVs in this study are controlled by SUMO's LC2013 model, a rule-based and heuristic decision policy. It is possible to learn human-like lane-changing decisions from empirical data and implement them to explore collective rationality in more realistic settings. Lastly, this study assumes that all AVs belong to the same company. However, different decision-making policies from various manufacturers can result in diverse AV behaviors. Therefore, it is worthwhile to explore collective rationality when multiple AV classes are considered using both analytical and learning-based methods.

\section*{Acknowledgment}{%
This research is supported by NSF grant ``EAGER: Collaborative Research: Fostering Collective Rationality Among Self-Interested Agents to Improve Design and Efficiency of Mixed Autonomy Networks and Infrastructure Systems'' (Award numbers: 2437983, 2437982).
}%

\bibliographystyle{plainnat} 
\bibliography{references}  






\end{document}